\documentclass{ecai} 

\usepackage{latexsym}
\usepackage{amssymb}
\usepackage{amsmath}
\usepackage{amsthm}
\usepackage{booktabs}
\usepackage{enumitem}
\usepackage{graphicx}
\usepackage{color}
\usepackage{acronym}
\usepackage{todonotes}
\usepackage{subfigure}

\usepackage{booktabs}
\usetikzlibrary{tikzmark}
\usepackage{xr}
\pgfdeclarelayer{bg}
\pgfsetlayers{bg,main}

\tikzset{
	rv/.style={draw, ellipse},
	pf/.style={draw, rectangle, fill = gray!30},
	arc/.style = {->, >={[round,sep]Stealth}},
}

\newcommand\factor[6]{
	\node[pf, #1=#3 of #2, label={#4:{#5}}](#6) {};
}

\usepackage{algorithm}
\usepackage[noend]{algpseudocode}
\algnewcommand{\algorithmicbreak}{\textbf{break}}
\algnewcommand{\algorithmiccontinue}{\textbf{continue}}
\algnewcommand{\algorithmicforeach}{\textbf{for each}}
\algnewcommand\Break{\State \algorithmicbreak}
\algnewcommand\Continue{\State \algorithmiccontinue}
\algdef{SE}[FOR]{ForEach}{EndForEach}[1]
	{\algorithmicforeach\ #1\ \algorithmicdo}
	{\algorithmicend\ \algorithmicforeach}
\algnewcommand{\LeftComment}[1]{\Statex \(\triangleright\) #1}
\algrenewcommand{\algorithmicindent}{1.15em}
\newcommand{\alginput}[1]{\hspace*{\algorithmicindent} \textbf{Input:} #1}
\newcommand{\algoutput}[1]{\hspace*{\algorithmicindent} \textbf{Output:} #1}

\usepackage[capitalise,]{cleveref}

\usepackage{tikz}
\newtheorem{example}{Example}
\crefname{algocf}{Alg.}{Algs.}
\Crefname{algocf}{Algorithm}{Algorithms}
\crefname{algorithm}{Alg.}{Algs.}
\Crefname{algorithm}{Algorithm}{Algorithms}
\crefname{definition}{Def.}{Defs.}
\Crefname{definition}{Definition}{Definitions}

\newtheorem{theorem}{Theorem}

\newtheorem{corollary}[theorem]{Corollary}
\newtheorem{proposition}[theorem]{Proposition}

\newtheorem{definition}{Definition}
\newtheorem{note}{Note}
\acrodef{rv}[randvar]{random variable} 
\acrodef{acp}[ACP]{Advanced Colour Passing}
\acrodef{bn}[BN]{Bayesian network}
\acrodef{cp}[CP]{Colour Passing}
\acrodef{fg}[FG]{factor graph}
\acrodef{eacp}[$\varepsilon$-ACP]{$\varepsilon$-Advanced Colour Passing}
\acrodef{hacp}[HACP]{Hierarchical Advanced Colour Passing}
\acrodef{odeed}[1DEED]{one-dimensional $\varepsilon$-equivalence distance}

\DeclareMathOperator*{\argmin}{arg\,min}

\begin{document}

\begin{frontmatter}

\paperid{7707} 

\title{Compression versus Accuracy:\\A Hierarchy of Lifted Models}


\author[A]{\fnms{Jan}~\snm{Speller}\orcid{0000-0003-1106-8177}\thanks{Corresponding Author. Email: jan.speller@uni-muenster.de}}
\author[B,C]{\fnms{Malte}~\snm{Luttermann}\orcid{0009-0005-8591-6839
}}
\author[C]{\fnms{Marcel}~\snm{Gehrke}\orcid{0000-0001-9056-7673}} 
\author[A]{\fnms{Tanya}~\snm{Braun}\orcid{0000-0003-0282-4284}} 

\address[A]{Computer Science Department, University of Münster, Germany}
\address[B]{German Research Center for Artificial Intelligence (DFKI), Lübeck, Germany}
\address[C]{Institute for Humanities-Centered Artificial Intelligence, University of Hamburg, Germany}

\begin{abstract}
Probabilistic graphical models that encode indistinguishable objects and relations among them use first-order logic constructs to compress a propositional factorised model for more efficient (lifted) inference.
To obtain a lifted representation, the state-of-the-art algorithm Advanced Colour Passing (ACP) groups factors that represent matching distributions.
In an approximate version using $\varepsilon$ as a hyperparameter, factors are grouped that differ by a factor of at most $(1\pm \varepsilon)$.
However, finding a suitable $\varepsilon$ is not obvious and may need a lot of exploration, possibly requiring many ACP runs with different $\varepsilon$ values.
Additionally, varying $\varepsilon$ can yield wildly different models, leading to decreased interpretability.
Therefore, this paper presents a hierarchical approach to lifted model construction that is hyperparameter-free.
It efficiently computes a hierarchy of $\varepsilon$ values that ensures a hierarchy of models, meaning that once factors are grouped together given some $\varepsilon$, these factors will be grouped together for larger $\varepsilon$ as well.
The hierarchy of $\varepsilon$ values also leads to a hierarchy of error bounds.
This allows for explicitly weighing compression versus accuracy when choosing specific $\varepsilon$ values to run ACP with and enables interpretability between the different models.
\end{abstract}

\end{frontmatter}

\section{Introduction}\label{sec:introduction}
Probabilistic graphical models (PGMs) allow for modelling environments under uncertainty by encoding features in \acp{rv} and relations between them in factors.
Lifted or first-order versions of PGMs such as parametric \acp{fg} \cite{Poo03} and Markov logic networks \cite{RicDo06} incorporate logic constructs to encode indistinguishable objects and relations among them in a compact way.
Probabilistic inference on such first-order models is tractable in the domain size if using representatives for indistinguishable objects \cite{NieBr14}, a technique referred to as lifting \cite{Poo03}.
Lifting has been used to great effect in probabilistic inference including lifting various query answering algorithms \cite{AhmKeMlNa13,BraMo16a,GogDo11,HarMoBr24,TagFiDaBl13,BroTaMeDaRa11} next to lifting queries \cite{BraMo18b,Bro13} or evidence \cite{TagFiDaBl13,BroDa12}.

\ac{acp} is the state-of-the-art algorithm to get a first-order model from a propositional one, specifically turning \acp{fg} into parametric \acp{fg}, grouping factors with identical potentials \cite{AhmKeMlNa13,LutBrMoGe24}.
The newest version, \ac{eacp}, considers approximate indistinguishability, using a number $\varepsilon$ to group factors whose potentials differ by a factor of at most $(1 \pm \varepsilon)$ and are therefore considered $\varepsilon$-equivalent, with $\varepsilon$ as a hyperparameter \cite{Luttermann2025a}.
\ac{eacp} allows for compressing a propositional model to a larger degree with increasing $\varepsilon$, leading to better runtime for inference tasks.
It also allows for computing an approximation error for such inference tasks, which helps assess the accuracy versus the compression gained. 
However, if a chosen $\varepsilon$ does not fulfil either a requirement for compression or accuracy, shortcomings become apparent:
A new run of \ac{eacp} with a different $\varepsilon$ does not guarantee a model consistent with the previous one.
E.g., with a larger $\varepsilon$, making more factors $\varepsilon$-equivalent, factors that were previously grouped together might no longer be part of the same group, because a different grouping appears more suitable.
That is, the models do not form a hierarchy, where groups of factors under a larger $\varepsilon$ can only form by merging groups under a smaller $\varepsilon$.
This inconsistency in models from one $\varepsilon$ to the next makes it hard to interpret the models regarding each other.
Additionally, $\varepsilon$ is a hyperparameter that has to be chosen by the user.
It may not be obvious what a suitable value for $\varepsilon$ is, requiring many runs of \ac{eacp} to find a suitable one with the necessary compression and accuracy.

To counteract these shortcomings, this paper presents a hierarchical approach to lifted model construction called hierarchical \ac{acp} (\acs{hacp})\acused{hacp}, which is hyperparameter-free, i.e., there is no need to choose a value for $\varepsilon$ in advance.
Specifically, we calculate a hierarchy of $\varepsilon$ values that guarantees a hierarchy of compressed models when running \ac{hacp}, as more compressed models naturally encompass supersets of grouped $\varepsilon$-equivalent factors from their less compressed counterparts.
To do so, this paper contributes the following: 
\begin{enumerate}[label=(\roman*)]
    \item a \ac{odeed} measure that reduces the $\varepsilon$-equivalence after computations to a single number for comparing different factors in contrast to the previous definition of $\varepsilon$-equivalence, 
    \item an efficient algorithm for computing a hierarchical ordering of $\varepsilon$ values,
    \item \ac{hacp} using the hierarchical ordering of $\varepsilon$ values, yielding a hierarchy of parametric \acp{fg}, and
    \item an analysis of the error bounds in the hierarchy of parametric \acp{fg}, showing a hierarchical order as well.
\end{enumerate}
The hierarchical approach has the advantage of being hyper\-pa\-ram\-e\-ter-free:
The hierarchy of $\varepsilon$ values can be computed before running \ac{hacp} and without needing a starting value.
In combination with the error bounds, the hierarchy of $\varepsilon$ values allows to choose for which $\varepsilon$ values to actually run \ac{hacp}, which means that a user does not need to do a hyperparameter exploration to find the most suitable one.
It also has the upside that one has to run \ac{hacp} only for those $\varepsilon$ values that are actually of interest. We approach this from the perspective of distributional deviation (accuracy), but also in the context of group merging of $\varepsilon$-equivalent factors by controlling $\varepsilon$ (compression).
In this context, we also investigate the potential loss of accuracy by the forced hierarchical structure in comparison to the \ac{eacp}.
Finally, the different models as a result of the $\varepsilon$ values in the hierarchy are consistent and as such interpretable with respect to each other.
This provides insight into the underlying symmetries within the different levels of the approximated \ac{fg}, where its complexity is implicitly captured by the variations and proximity of distinct $\varepsilon$ values and the heterogeneity of group memberships.
Consequently, this allows for an informed choice of an appropriate $\varepsilon$ in a way that suits specific requirements for applications.

The remaining part of this paper is structured as follows:
The paper starts with introducing necessary notations and briefly recaps \ac{eacp}, which is followed by the main part, which provides a definition of \ac{odeed}, specifies how to calculate a hierarchical ordering of $\varepsilon$ values, and presents \ac{hacp}.
Then, it shows maximal error bounds for \ac{hacp} and ends with a discussion and conclusion. 
The technical appendix includes more detailed proofs and illustrations.

\section{Background}\label{sec:background}
We start by defining an \ac{fg} as a propositional probabilistic graphical model to compactly encode a full joint probability distribution over a set of \acp{rv}~\citep{Frey1997a,Kschischang2001a}.
The following definitions are given via \cite{Luttermann2025a}.

\begin{definition}[Factor Graph] \label{def:fg}
	An \ac{fg} $M = (\boldsymbol V, \boldsymbol E)$ is an undirected bipartite graph consisting of a node set $\boldsymbol V = \boldsymbol R \cup \boldsymbol \Phi$, where $\boldsymbol R = \{R_1, \ldots, R_n\}$ is a set of \acp{rv} and $\boldsymbol \Phi = \{\phi_1, \ldots, \phi_m\}$ is a set of factors (functions), as well as a set of edges $\boldsymbol E \subseteq \boldsymbol R \times \boldsymbol \Phi$.
	There is an edge between a \ac{rv} $R_i \in \boldsymbol R$ and a factor $\phi_j \in \boldsymbol \Phi$ in $\boldsymbol E$ if $R_i$ appears in the argument list of $\phi_j$.
	A factor $\phi_j(\mathcal R_j)$ defines a function $\phi_j \colon \times_{R \in \mathcal R_j} \text{range}(R) \to \mathbb{R}_{>0}$ that maps the ranges of its arguments $\mathcal R_j$ (a sequence of \acp{rv} from $\boldsymbol R$) to a positive real number, called potential.
	The term $\text{range}(R)$ denotes the possible values a \ac{rv} $R$ can take.
	We further define the joint potential for an assignment $\boldsymbol r$ (with $\boldsymbol r$ being a shorthand notation for $\boldsymbol R = \boldsymbol r$) as
	\begin{align}
		\psi(\boldsymbol r) = \prod_{j=1}^m \phi_j(\boldsymbol r_j),
	\end{align}
	where $\boldsymbol r_j$ is a projection of $\boldsymbol r$ to the argument list of $\phi_j$.
	With $Z = \sum_{\boldsymbol r} \prod_{j=1}^{m} \phi_j(\boldsymbol r_j)$ as the normalisation constant, the full joint probability distribution encoded by $M$ is then given by 
	\begin{align}
		P_M(\boldsymbol r) = \frac{1}{Z} \prod_{j=1}^m \phi_j(\boldsymbol r_j) = \frac{1}{Z} \psi(\boldsymbol r).
	\end{align}
\end{definition}

\begin{figure}[b]
    \centering
    \begin{tikzpicture}[scale=0.9]
    \node[circle, draw] (A) {$A$};
	\node[circle, draw] (B) [below = 0.25cm of A] {$B$};
	\node[circle, draw] (C) [below = 0.25cm of B] {$C$};
	\factor{below right}{A}{0.125cm and 0.5cm}{270}{$\phi_1$}{f1}
	\factor{below right}{B}{0.125cm and 0.5cm}{270}{$\phi_2$}{f2}
\node[scale=0.9 ,right = 0.125cm of f1, yshift=-4mm] (tabs) {
    \begin{tabular}{c|c|c c@{\hspace{0.25cm}}c|c|c}
        $A$ & $B$ & $\phi_1(A,B)$ & & $C$ & $B$ & $\phi_2(C,B)$ \\ \cline{1-3} \cline{5-7}
        true  & true  & $\varphi_1$ & & true  & true  & $\varphi_1$ \\
        true  & false & $\varphi_2$ & & true  & false & $\varphi_2$ \\
        false & true  & $\varphi_3$ & & false & true  & $\varphi_3$ \\
        false & false & $\varphi_4$ & & false & false & $\varphi_4$ \\
    \end{tabular}
};

	\draw (A) -- (f1);
	\draw (B) -- (f1);
	\draw (B) -- (f2);
	\draw (C) -- (f2);
\end{tikzpicture}
    \caption{An exemplary \ac{fg} encoding a full joint probability distribution over three \acp{rv} $A$, $B$, and $C$.}
    \label{fig:example_fg}
\end{figure}
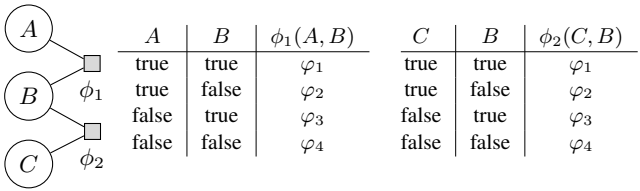

\begin{example}
    Consider the \ac{fg} illustrated in \cref{fig:example_fg}.
    It holds that $\boldsymbol R = \{A, \allowbreak B, \allowbreak C\}$, $\boldsymbol \Phi = \{\phi_1, \allowbreak \phi_2\}$, and $\boldsymbol E = \{ \{A, \allowbreak \phi_1\}, \allowbreak \{B, \allowbreak \phi_1\}, \allowbreak \{B, \allowbreak \phi_2\}, \allowbreak \{C, \allowbreak \phi_2\} \}$.
	For the sake of this example, let $\text{range}(A) = \allowbreak \text{range}(B) = \allowbreak \text{range}(C) = \allowbreak \{\text{\emph{true}}, \allowbreak \text{\emph{false}}\}$.
	The potential tables of $\phi_1$ and $\phi_2$ are shown on the right of \cref{fig:example_fg} with $\phi_1(\text{\emph{true}}, \text{\emph{true}}) = \varphi_1$ and so on, where $\varphi_i \in \mathbb{R}_{>0}$, $i = 1, \allowbreak \ldots, \allowbreak 4$, are arbitrary positive real numbers.
\end{example}

In \cref{def:fg}, we stipulate that all potentials are strictly greater than zero to avoid division by zero when analysing theoretical bounds in subsequent sections of this paper.
In general, it is sufficient to have at least one non-zero potential in every potential table to ensure a well-defined semantics of an \ac{fg}.
However, our requirement of having strictly positive potentials is no restriction in practice as zeros can easily be replaced by tiny numbers that are close to zero.
An \ac{fg} can be queried to compute marginal distributions of \acp{rv} given observations for other \acp{rv} (referred to as probabilistic inference).

\begin{definition}[Query]
	A \emph{query} $P(Q \mid E_1 = e_1, \ldots, E_k = e_k)$ consists of a query term $Q$ and a set of events $\{E_j = e_j\}_{j=1}^{k}$ where $Q$ and all $E_j$, $j = 1, \allowbreak \ldots, \allowbreak k$, are \acp{rv}.
	To query a specific probability instead of a distribution, the query term is an event $Q = q$.
\end{definition}

Lifted inference exploits identical behaviour of indistinguishable objects to answer queries more efficiently.
The idea behind lifting is to use a representative of indistinguishable objects for computations.
Formally, this corresponds to making use of exponentiation instead of multiplying identical potentials several times (for an example, see Appendix~3). 
To exploit exponentiation during inference, equivalent factors have to be grouped.
We next introduce the concept of $\varepsilon$-equivalent factors~\citep{Luttermann2025a}, which allows us to determine factors that can potentially be grouped for lifted inference.

\begin{definition}[$\varepsilon$-Equivalence] \label{def:original_epsilon_equivalence}
	Let $\varepsilon \in \mathbb{R}_{>0}$ be a positive real number.
	Two potentials $\varphi_1,\varphi_2 \in \mathbb{R}_{>0}$ are \emph{$\varepsilon$-equivalent}, denoted as $\varphi_1 =_{\varepsilon} \varphi_2$, if $\varphi_1 \in [\varphi_2 \cdot (1 - \varepsilon), \varphi_2 \cdot (1 + \varepsilon)]$ and $\varphi_2 \in [\varphi_1 \cdot (1 - \varepsilon), \varphi_1 \cdot (1 + \varepsilon)]$.
	Further, two factors $\phi_1(R_1, \ldots, R_n)$ and $\phi_2(R'_1, \ldots, R'_n)$ are \emph{$\varepsilon$-equivalent}, denoted as $\phi_1 =_{\varepsilon}\phi_2$, if there exists a permutation $\pi$ of $\{1, \ldots, n\}$ such that for all assignments $(r_1, \ldots, r_n) \in \times_{i=1}^n \text{range}(R_i)$, where $\phi_1(r_1, \ldots, r_n) = \varphi_1$ and $\phi_2(r_{\pi(1)}, \ldots, r_{\pi(n)}) = \varphi_2$, it holds that $\varphi_1 =_{\varepsilon} \varphi_2$.
\end{definition}

\begin{example}
	Consider the potentials $\varphi_1 = 0.49$, $\varphi_2 = 0.5$, and $\varepsilon = 0.1$.
	Since it holds that $\varphi_2 = 0.5 \in [\varphi_1 \cdot (1 - \varepsilon) = 0.441, \varphi_1 \cdot (1 + \varepsilon) = 0.539]$ and $\varphi_1 = 0.49 \in [\varphi_2 \cdot (1 - \varepsilon) = 0.45, \varphi_2 \cdot (1 + \varepsilon) = 0.55]$, $\varphi_1$ and $\varphi_2$ are $\varepsilon$-equivalent (for $\varepsilon=0.1)$.
\end{example}

The notion of $\varepsilon$-equivalence is symmetric.
Moreover, it might happen that indistinguishable objects are located at different positions in the argument list of their respective factors, which is the reason the definition considers permutations of arguments.
For simplicity, in this paper, we stipulate that $\pi$ is the identity function.
However, all presented results also apply to any other choice of $\pi$~\citep{Luttermann2025a}.

The \ac{eacp} algorithm~\citep{Luttermann2025a} computes groups of pairwise $\varepsilon$-equivalent factors to compress a given \ac{fg}.
In particular, as potentials are often estimated in practice, potentials that should actually be considered equal might slightly differ and the \ac{eacp} algorithm accounts for such deviations using a hyperparameter $\varepsilon$, which controls the trade-off between compression and accuracy of the resulting lifted representation.
To allow for exponentiation, \ac{eacp} computes the mean potentials for each group of pairwise $\varepsilon$-equivalent factors and replaces the original potentials of the factors by the respective mean potentials.
Thus, the semantics of the \ac{fg} changes after the replacement of potentials.
However, replacing potentials by their mean guarantees specific error bounds (more details follow in \cref{sec:bounded}).

\begin{note}
    \Ac{eacp} uses the introduced concept of $\varepsilon$-equivalence including the corresponding hyperparameter $\varepsilon\in \mathbb{R}_{>0}$ requiring repeated, dimension-wise grouping checks with no reusable aggregate structure. 
    This can be interpreted as a regularisation approach from the original model (with $\varepsilon\approx 0$) to the most trivial model by reducing complexity by increasing $\varepsilon$. 
    For arbitrary large $\varepsilon \gg 0$, it reduces the model to an \ac{fg} that considers all factors of the same dimension and its corresponding multiset of \acp{rv} with same range sizes as pairwise $\varepsilon$-equivalent, where the dimension of a factor $\phi$ refers to the number of rows in the potential table of $\phi$.
    \Ac{eacp} does not yield hierarchical models, since the grouping process is independent for each choice of $\varepsilon$.
\end{note} 
Next, we present a hierarchical version of \ac{eacp} that ensures a hierarchy of models over increasing $\varepsilon$.

\section{Hierarchical Lifted Model Construction}\label{sec:hierarchical}
As mentioned above, \ac{eacp} lacks a mechanism to ensure the core property of hierarchical methods: the consistent embedding of simpler models into more complex ones.
To construct a principled hierarchical organisation of models, a clear mechanism for defining and transferring group membership across levels is essential. 
Concretely, we define a hierarchy, in which higher levels inherit structural properties from lower levels, thereby inducing a consistent reduction in the complexity of the \ac{fg}.
To this end, we present the \emph{\ac{odeed}}, a more effective criterion for determining $\varepsilon$-equivalence. 
To do so, we treat the potential table of a factor $\phi$ as a vector in $\mathbb{R}_{>0}^n$, where $\phi(k)$ denotes the $k$-th entry, i.e., the potential associated with the $k$-th row in the potential table of $\phi$.
For example, factor $\phi_1(A,B)$ in \cref{fig:example_fg} is represented as the vector $(\varphi_1,\varphi_2,\varphi_3,\varphi_4)$, with, e.g., $\phi_1(\text{true},\text{false})=\phi_1(2) = \varphi_2$.
After introducing \ac{odeed}, we use it to set up a hierarchical ordering of $\varepsilon$-equivalent factors, which forms the backbone to a hierarchical approach for lifted model construction based on \ac{eacp}.

\subsection{One-dimensional $\varepsilon$-Equivalence Distance}\label{subsec:one_dimensional}
We define \ac{odeed} as a measure to compare two $n$-dimensional, strictly positive vectors, representing factors in an \ac{fg}, i.e., $\phi_i \in \mathbb{R}_{>0}^n$ with $\phi_i(k) > 0$ for all $k$.

\begin{definition}[One-dimensional $\varepsilon$-equivalence distance] \label{def:one_dimensional_epsilon_equivalence}
    \emph{\ac{odeed}}
    defined as the mapping $d_{\infty}\colon \mathbb{R}_{>0}^n \times \mathbb{R}_{>0}^n \to \mathbb{R}$ 
    for two n-dimensional vectors $\phi_1,\phi_2\in\mathbb{R}_{>0}^n$ is given by:
    \begin{align}
        d_{\infty}(\phi_1,\phi_2)&:= \max_{k=1,\ldots,n} \left\{\left\vert  \frac{\phi_1(k)-\phi_2(k)}{\phi_1(k)}\right\vert,\left\vert \frac{\phi_1(k)-\phi_2(k)}{\phi_2(k)}\right\vert\right\}\nonumber\\
        &~= \max_{k=1,\ldots,n} \left\{ \frac{\vert\phi_1(k)-\phi_2(k)\vert}{\min\{\vert\phi_1(k)\vert,\vert\phi_2(k)\vert\}}\right\} \label{eq:odeed}
    \end{align}
\end{definition}
The first properties of \ac{odeed} as a distance measure are direct results of its definition.
\begin{corollary}\label{corellary:positiveandsymmetric}
The following properties hold for \ac{odeed}.
\begin{itemize}
    \item [(i)] \Ac{odeed} is non-negative and symmetric.
    \item [(ii)] It holds that $d_{\infty}(\phi_1,\phi_2)=0$ if and only if $\vert\phi_1(k)-\phi_2(k)\vert = 0 $ for all $k =1,\ldots,n$, which holds if and only if $\phi_1=\phi_2$.
\end{itemize}
\end{corollary}

This distance is based on the \textit{maximum metric (Chebyshev distance)} with an additional deviation. 
It does not satisfy the triangle inequality and thus is not a metric, which can be demonstrated using a counterexample (see Appendix~1). 
Nonetheless, \ac{odeed} is well-suited for bounding maximum relative deviations in all dimensions, aligning with practical use in probabilistic inference.
The nature of its definition is no coincidence, but rather subject to the purpose to induce $\varepsilon$-equivalence by being consistent with the existing concept. 

\begin{theorem}\label{theorem:equivalent_varepsilon_definitions}
Two vectors $\phi_1,\phi_2\in\mathbb{R}_{>0}^n$ are $\varepsilon$-equivalent (\Cref{def:original_epsilon_equivalence}) if and only if $d_{\infty}(\phi_1,\phi_2)\leq \varepsilon$ holds.
\end{theorem}
\begin{proof}[Proof Sketch]
     In mathematical terms, the claim can be summarised as $\phi_1 =_{\varepsilon}\phi_2 \Leftrightarrow d_{\infty}(\phi_1,\phi_2)\leq \varepsilon$, which we prove for any $\varepsilon>0$ in Appendix~2 via equivalence transformations.\end{proof}

The use of relative deviation is essential in probabilistic querying, where percentage-based error bounds directly influence the quality of inference. 
However, classical definitions of $\varepsilon$-equivalence require checking all component-wise comparisons, leading to inefficiencies in practice and no clear order of $\varepsilon$-equivalent factor groups.
The one-dimensional formulation via $d_{\infty}$ addresses this by providing a closed-form, computationally minimal characterisation of $\varepsilon$-equivalence. 
Specifically, it allows for an efficient computation of the smallest admissible $\varepsilon$ for each pair of factors, facilitating both storage and comparison.
That is $d_{\infty}(\phi_1,\phi_2)=\varepsilon_0$ uniquely determines the minimal $\varepsilon_0\leq \varepsilon$ for which $\phi_1 =_\varepsilon \phi_2$ still holds. 
This scalar value enables direct comparison, indexing, and efficient classification of equivalence classes without enumerating all component-wise ratios. 
The formal operational benefits -- both in terms of computational complexity and structural hierarchy -- are established in the next subsection.

\subsection{Hierarchical Ordering of $\varepsilon$-Equivalent Factors}\label{subsec:preorder}

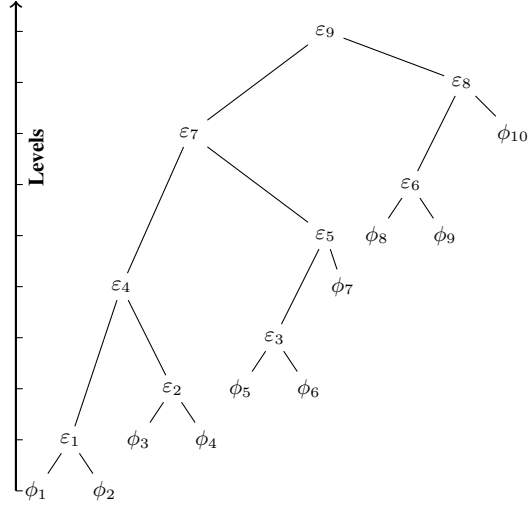
\begin{figure}[t]
    \centering
    \begin{tikzpicture}[scale=0.9, transform shape, node distance=1.2cm] 
        \node (phi1) at (0,0) {$\phi_1$};
        \node (phi2) at (1.0,0) {$\phi_2$};
        \node (12) at (0.5,0.75) {$\varepsilon_1$};
        \node (phi3) at (1.5,0.75) {$\phi_3$};
        \node (phi4) at (2.5,0.75) {$\phi_4$};
        \node (34) at (2.0,1.5) {$\varepsilon_2$};
        \node (phi5) at (3.0,1.5) {$\phi_5$};
        \node (phi6) at (4.0,1.5) {$\phi_6$};
        \node (56) at (3.5,2.25) {$\varepsilon_3$};
        \node (1234) at (1.25,3.0) {$\varepsilon_4$};
        \node (phi7) at (4.5,3.0) {$\phi_7$};
        \node (567) at (4.25,3.75) {$\varepsilon_5$};
        \node (root) at (2.25,5.25) {$\varepsilon_7$};
        \node (phi8) at (5.0,3.75) {$\phi_8$};
        \node (phi9) at (6.0,3.75) {$\phi_9$};
        \node (89) at (5.5,4.5) {$\varepsilon_6$};
        \node (phi10) at (7.0,5.25) {$\phi_{10}$};
        \node (phi8910) at (6.25,6.0) {$\varepsilon_8$};
        \node (phirest8910) at (4.25,6.75) {$\varepsilon_9$};
        
        \draw (phi1) -- (12);
        \draw (phi2) -- (12);
        \draw (12) -- (1234);
        \draw (phi3) -- (34);
        \draw (phi4) -- (34);
        \draw (34) -- (1234);
        \draw (phi5) -- (56);
        \draw (phi6) -- (56);
        \draw (56) -- (567);
        \draw (phi7) -- (567);
        \draw (1234) -- (root);
        \draw (567) -- (root);
        \draw (phi8) -- (89);
        \draw (phi9) -- (89);
        \draw (89) -- (phi8910);
        \draw (phi10) -- (phi8910);
        \draw (phi8910) -- (phirest8910);
        \draw (root) -- (phirest8910);
        \draw[->, thick] (-0.3,0.0) -- (-0.3,7.2);
        \node[rotate=90] at (-0.0,4.9) {\textbf{Levels}};
        \foreach \y in {0.0,0.75,...,7.0} {
            \draw (-0.3,\y) -- (-0.2,\y);
        }
    \end{tikzpicture}
    \caption{Exemplary visualisation of a factor ordering with increasing $\varepsilon$. This information is easily stored in the list $\mathcal{L}$ and is easily readable from the matrix $\tilde{\Lambda}$. The $\varepsilon_i$ are ordered by size ($\varepsilon_1$ being the smallest value of them). Note that a root $\varepsilon$ is always the maximal $\varepsilon$-distance of all pairwise factor comparisons of all leafs. E.g., $\varepsilon_4 = \max\{\varepsilon_{1,2},\varepsilon_{1,3},\varepsilon_{1,4},\varepsilon_{2,3},\varepsilon_{2,4},\varepsilon_{3,4}\}$.}\label{fig:hierarchy}
\end{figure}

To obtain a meaningful hierarchical structure, we need a starting point or a level $0$, which is simply the full model. 
On the other hand, if $\varepsilon$ is arbitrarily large, all factors of the same dimension would be grouped together, resulting in a nearly trivial model. 
However, these properties alone are insufficient for creating a clear hierarchy between these extremes. 
A desired structure would resemble the one depicted in \cref{fig:hierarchy}.
Each level in the hierarchy is represented by a line on the left side of the figure. 
If we were to cut the figure horizontally at this point, all connected subtrees would form a group, while the remaining factors would stay separate. 
Our goal is to maintain the property that once multiple factors are grouped together at a lower level, they should also stay together at higher levels. 
This property is known as \text{structure preservation}. 
However, preserving the structure is not trivial due to the lack of transitivity in $d_{\infty}$.

Thus, we impose a hierarchical pre-ordering on the set of factors based on pairwise $\varepsilon$-equivalence, determined by the minimal deviation under \ac{odeed} $d_{\infty}$. 
This unique ordering forms the basis for \ac{hacp} described in \cref{subsection:HACP}. 
Smaller $d_{\infty}$ values correspond to higher indistinguishability, and the hierarchical construction ensures that higher-level aggregations preserve the nested structure of previously merged subgroups.
This can be seen as an agglomerative clustering algorithm based on \ac{odeed} with complete linkage within maximal deviation (where only completely pairwise $\varepsilon$-equivalent groups are merged) with complexity $O(m^3)$, which is especially in comparison to the complexity of \ac{acp} negligible. The grouping step of \ac{eacp} is omitted instead, which is applied per choice of $\varepsilon$. 

To determine this ordering, we follow a two-phase procedure as described in \cref{alg:ordering} with an \ac{fg} as input.
In Phase~I, the algorithm creates a matrix $\Lambda$ as shown in \cref{tab:matrix_updating_system}. 
Its cell entries are $\Lambda_{ij} := \varepsilon_{i,j} := d_{\infty}(\phi_i, \phi_j)$ for $1 \leq  i < j \leq m$, where $m = |\boldsymbol{\Phi}|$ is the number of factors in the \ac{fg}. 
The symmetric property of $d_{\infty}$
allows for filling the matrix with zeros, thus forming an upper triangular matrix. 
Next, we examine Phase~II of the algorithm in more detail, which performs a hierarchical ordering of $\varepsilon$-equivalent group selections across all factors with structural compatibility.

\begin{table}[t]
    \centering
    \begin{tikzpicture}
	\node (t1) {
		\setlength{\tabcolsep}{5pt}
		\begin{tabular}{c|cccccc}
			\toprule
Factor & $\phi_1$ & $\phi_2$ & $\cdots$ &$\cdots$  & $\phi_{m-1}$ & $\phi_m$ \\ \midrule
			$\phi_1$ & 0 & $\varepsilon_{1,2}$ & $\varepsilon_{1,3}$ & $\cdots$ & $\varepsilon_{1,m-1}$& $\varepsilon_{1,m}$\\
$\phi_2$& 0& 0& $\varepsilon_{2,3}$& $\cdots$ & $\varepsilon_{2,m-1}$& $\varepsilon_{2,m}$ \\
$\vdots$& $\vdots$& $\vdots$& $\ddots$& $\ddots$ & $\vdots$ & $\vdots$\\
$\vdots$& $\vdots$& $\vdots$& $\vdots$& $\ddots$&$\ddots$  & $\vdots$\\
$\phi_{m-1}$& 0& 0& 0& 0 & 0 & $\varepsilon_{m-1,m}$\\
$\phi_m$ & 0&0 &0 &0 &0 &0  \\
            \bottomrule  
		\end{tabular}
	};
\end{tikzpicture}
    \caption{Upper triangular matrix $\Lambda = (\Lambda_{ij})_{1 \leq i , j \leq m}$ illustrating the matrix output of \cref{alg:ordering}, Phase I. The entries are defined as $\Lambda_{ij} = \varepsilon_{i,j} := d_{\infty}(\phi_i, \phi_j)$ for $1 \leq  i < j \leq m$, and $\Lambda_{ij} = 0$ otherwise.}
    \label{tab:matrix_updating_system}
\end{table}

\begin{algorithm}[t]
	\caption{Hierarchical Ordering of $\varepsilon$-Equivalent Groups}\label{alg:ordering}
    \alginput{An \ac{fg} $M = (\boldsymbol R \cup \boldsymbol \Phi, \boldsymbol E)$ with $m=|\boldsymbol{\Phi}|$ and $\boldsymbol \Phi\subset\mathbb{R}_{>0}^{n\times m}$,\\
    \hspace*{\algorithmicindent} and structural compatibility across all factors.}\\
    \algoutput{
    An ordered nested list $\mathcal{L}$ of lists,\\\hspace*{\algorithmicindent} 
    an ordered vector $\boldsymbol{\varepsilon}=(\varepsilon_1\ldots,\varepsilon_{m-1})$ with $\varepsilon_i<\varepsilon_{i+1}$.}\\
    \begin{algorithmic}[1]
        \LeftComment{Phase I: Generate upper triangular Matrix $\Lambda$}
        \State Initialise $\Lambda \in \mathbb{R}^{m \times m}$ with zeros
        \For{$i = 1$ to $m-1$}
            \For{$j = i+1$ to $m$}
                \State Compute $d_{\infty}(\phi_i, \phi_j)=:\varepsilon_{i,j}$
                \State Store result in $\Lambda_{ij}$
            \EndFor
        \EndFor
        \State Save $\Lambda$\hspace*{\algorithmicindent}
        \LeftComment{Phase II: Generate ordered list $\mathcal{L}$ and vector $\boldsymbol{\varepsilon}$}
        \State Initialise empty list $\mathcal{L} \gets [\,]$
        \State Initialise vector $\boldsymbol{\varepsilon}=(\varepsilon_1\ldots,\varepsilon_{m-1})$ with zeros
        \State Initialise active index set $\mathcal{A} \gets \{1, \ldots, m\}$
        \State Initialise active matrix $\tilde{\Lambda}:=\Lambda$
        \For{$\ell = 1$ to $m-1$}
            \State Find $(i', j') = \arg\min\{\tilde{\Lambda}_{ij} \mid i < j,\; i,j \in \mathcal{A}\}$
            \State Save $\varepsilon_l:=\tilde{\Lambda}_{i'j'}$
            \If{$i'$ appears in direct parent group $G_p \in\mathcal{L}$}
                \If{$j'$ appears in direct parent group $G_p' \in\mathcal{L}$} 
                    \State Update $G_p$ and $G_p'$ as one new group $[G_p,G_p',\ell+m]$
                \Else
                    \State Update $G_p$ as $G_p=[G_p,j',\ell+m]$ 
                \EndIf
            \ElsIf{$j'$ appears in direct parent group $G_p  \in\mathcal{L}$}
                \State Update $G_p$ as $G_p=[G_p,i',\ell+m]$
            \Else 
                \Comment Neither $i'$ nor $j'$ appears in any parent group
                \State Append $[[i', j', \ell+m]]$ to $\mathcal{L}$
            \EndIf
            \For{$k \in \mathcal{A} \setminus \{i',j'\}$}
                \If{$k > j'$}
                    \State $\tilde{\Lambda}_{i'k} \gets \max\{\tilde{\Lambda}_{i'k}, \tilde{\Lambda}_{j'k}\}$
                \ElsIf{$k < i'$}
                    \State $\tilde{\Lambda}_{ki'} \gets \max\{\tilde{\Lambda}_{ki'}, \tilde{\Lambda}_{kj'}\}$
                \ElsIf{$i' < k < j'$}
                    \State $\tilde{\Lambda}_{i'k} \gets \max\{\tilde{\Lambda}_{i'k}, \tilde{\Lambda}_{kj'}\}$
                \EndIf
            \EndFor
            \State Remove $j'$ from active set: $\mathcal{A} \gets \mathcal{A} \setminus \{j'\}$
        \EndFor 
	\end{algorithmic}
\end{algorithm}
The algorithm iteratively chooses $\varepsilon$ values from $\Lambda$ that allow (groups of) factors that are pairwise $\varepsilon$-equivalent to be grouped.
The algorithm runs for $m-1$ iterations as there are $m$ factors to merge, meaning there are $m-1$ hierarchical levels at the end.
The outputs are an ordered vector $\boldsymbol{\varepsilon}$ of length $m-1$ of increasing $\varepsilon$ values as well as an ordered nested list of lists $\mathcal{L}$ containing a nested grouping of indices according to the $\varepsilon$ values and their hierarchy level (with an index shift of $m$ for easier identification compared to the indices identifying factors).
Specifically, the algorithm picks the next two (groups of) factors to merge by selecting the minimal entry $\varepsilon_{i',j'}$ in $\Lambda$, which is then stored in $\boldsymbol{\varepsilon}$ at the current level.

Before dealing with $\mathcal{L}$, let us consider how $\Lambda$ is updated:
Since both (groups of) factors are now considered as a single group, their respective rows in $\Lambda$ need to be merged by keeping the maximum of the two $\varepsilon$ values in each column, which ensures that if an entry of this row is picked in another iteration, all factors are $\varepsilon$-equivalent given this larger $\varepsilon$.
To avoid resizing $\Lambda$, there is a set of active indices, and merging (groups of) factors removes the second index, essentially deactivating the row.
The entries of the row of the first index are then updated to the maximum value.

Regarding $\mathcal{L}$, a new entry is formed, which is essentially a list of a $3$-tuple $l =[e_{i'}, e_{j'}, h]$: one element $e_{i'}$ for the first (group of) factor(s), one element $e_{j'}$ for the second (group of) factor(s), and the last element $h$ being the current hierarchy level shifted by $m$.
If $i'$ or $j'$ identify a single factor, then $e_{i'}$ or $e_{j'}$ store the index identifying the factor.
If $i'$ or $j'$ identify a group of factors, then there already exists an entry $l'$ in $\mathcal{L}$ for it from a previous merging, which is removed from $\mathcal{L}$ and then stored in $e_{i'}$ or $e_{j'}$.
Next, we look at an example.

\begin{example}\label{example:alltogether}
Let $M = (\boldsymbol{R} \cup \boldsymbol{\Phi}, \boldsymbol{E})$ be an \ac{fg} with $m = |\boldsymbol{\Phi}| = 10$ factors, all of identical dimension. 
Assume distances between factors leading to a hierarchy corresponding to \cref{fig:hierarchy}, that is, factors $\phi_1$ and $\phi_2$ have the smallest distance among all factors, $\phi_3$ and $\phi_4$ the next smallest distance, followed by $\phi_5$ and $\phi_6$, after which the first two pairings have the smallest distance, and so on.

During the first iteration of Phase II in \cref{alg:ordering}, $\varepsilon_{1,2}$ is minimal in $\Lambda$.
Therefore, an entry in $\mathcal{L}$ is created with $[1,2,11]$, containing the two indices $1,2$ identifying the factors and the current hierarchy level $1$ shifted by $m=10$.
The matrix update looks as follows:
\begin{center}
    \begin{tikzpicture} 
\node {
\setlength{\tabcolsep}{5pt}
\begin{tabular}{c|*{5}{c}}
    \toprule
    Factor &  $\phi_2$ & $\phi_3$ &$\phi_4$ & $\phi_{5}$&$\cdots$ \\ \midrule
    $\phi_1$  & $\varepsilon_{1,2}$&$\displaystyle \max_{\substack{i=1,2 }}\varepsilon_{i,3} $ &$\displaystyle \max_{\substack{i=1,2 }}\varepsilon_{i,4} $  & $\displaystyle \max_{\substack{i=1,2 }}\varepsilon_{i,5} $& $\cdots$\\
    \tikzmark{start}$\phi_2$ & 0 & $\varepsilon_{2,3}$&$\varepsilon_{2,4}$& $\varepsilon_{2,5}$ & $\ldots$\tikzmark{end} \\
    $\phi_3$& 0& 0 &$\varepsilon_{3,4}$& $\varepsilon_{3,5}$ & $\cdots$\\
    $\phi_4$&  0 &0&0 &$\varepsilon_{4,5}$ &$\cdots$\\
    $\vdots$&  0& 0& 0 & 0 & $\cdots$\\
    \bottomrule  
\end{tabular}
};
\draw[red, thick] ([xshift=-100pt,yshift=4pt]pic cs:colstart) -- ([xshift=100pt,yshift=+4pt]pic cs:colend);
\draw[blue, thick] 
    ([xshift=-58pt,yshift=+36pt]pic cs:colstart) 
    -- 
    ([xshift=-58pt,yshift=-36pt]pic cs:colend);
\end{tikzpicture}
\end{center}
The row of index $1$ is updated to the maximum value of the two entries of the rows $1,2$.
The row of index $2$ is deactivated, which is depicted by the crossed out line.

In the next iteration, $\varepsilon_{3,4}$ is minimal in $\tilde{\Lambda}$, which means adding an entry $[3,4,12]$ to $\mathcal{L}$, with the matrix update being the following:
\begin{center}
    \begin{tikzpicture}
\node {
\setlength{\tabcolsep}{5pt}
\begin{tabular}{c|*{5}{c}}
    \toprule
    Factor &  $\phi_2$ & $\phi_3$ &$\phi_4$ & $\phi_{5}$&$\cdots$ \\ \midrule
    $\phi_1$  & $\varepsilon_{1,2}$&$\displaystyle \max_{\substack{i=1,2 \\ j=3,4}}\varepsilon_{i,j} $ &$\displaystyle \max_{\substack{i=1,2 }} \varepsilon_{i,4} $  & $\displaystyle \max_{\substack{i=1,2}} \varepsilon_{i,5} $& $\cdots$\\
    \tikzmark{start}$\phi_2$ & 0 & $\varepsilon_{2,3}$&$\varepsilon_{2,4}$& $\varepsilon_{2,5}$ & $\ldots$\tikzmark{end} \\
    $\phi_3$& 0& 0 &$\varepsilon_{3,4}$& $\displaystyle \max_{\substack{i=3,4 \\ }}\varepsilon_{i,5}$ & $\cdots$\\
    $\phi_4$&  0 & 0 & 0 &$\varepsilon_{4,5}$ &$\cdots$\\
    $\vdots$&  0& 0& 0 & 0 & $\cdots$\\
    \bottomrule  
\end{tabular}
};
\draw[red, thick] ([xshift=-100pt,yshift=3pt]pic cs:colstart) -- ([xshift=100pt,yshift=+3pt]pic cs:colend);
\draw[red, thick, dashed] ([xshift=-100pt,yshift=-23pt]pic cs:colstart) -- ([xshift=100pt,yshift=-23pt]pic cs:colend);
\draw[blue, thick] 
    ([xshift=-58pt,yshift=+40pt]pic cs:colstart) 
    -- 
    ([xshift=-58pt,yshift=-40pt]pic cs:colend);
\draw[blue, thick, dashed] 
    ([xshift=+18pt,yshift=+40pt]pic cs:colstart) 
    -- 
    ([xshift=+18pt,yshift=-40pt]pic cs:colend);
\end{tikzpicture}
\end{center}

When $\tilde{\Lambda}_{13}$ is the next value to choose, both indices identify groups of factors.
As such, their entries in $\mathcal{L}$, namely, $[1,2,11]$ and $[3,4,12]$, are replaced by an entry $[[1,2,11],[3,4,12],14]$.
At the end, the output of \cref{alg:ordering} looks as follows:
\begin{align*}
    \mathcal{L}&= [~[[[1,2, 11],[3,4,12],14],[[5,6,13],7,15],17],\\
    &\quad~~[[8,9,16], {10},18],19~]\\
    \boldsymbol{\varepsilon} &= (\varepsilon_1,\ldots,\varepsilon_{9}) \text{ with } ~~~~~~~~~~\varepsilon_1 = \min\limits_{\substack{i,j=1,\ldots,10\\i< j}}\{\varepsilon_{i,j}\} 
 = \Lambda_{12}, \\
 \varepsilon_2 &= \min\limits_{\substack{i,j=1,3,\ldots,10\\i< j}}\{\varepsilon_{i,j}\} 
= \tilde{\Lambda}_{34},~~\varepsilon_3 = \min\limits_{\substack{i,j=1,3,5,\ldots,10\\i< j}}\{\varepsilon_{i,j}\} 
= \tilde{\Lambda}_{56},\\
\varepsilon_4 &= \min\limits_{\substack{i,j=1,3,5,7,\ldots,10\\i< j}}\{\varepsilon_{i,j}\} 
= \tilde{\Lambda}_{13},~~~~\ldots
    \end{align*}

This results in a total hierarchy of maximal $10$ different levels (and models).
Appendix~4 shows an overview of the group sizes in the hierarchy, illustrating the compression possible with increasing $\varepsilon$. 
\end{example}

Thus, in the output, each $\varepsilon_i$ 
corresponds to an increasingly coarse partitioning, reflecting group memberships under growing tolerance thresholds for higher levels.
Selecting a specific $\varepsilon_i$ implies fixing a hierarchical level $i$, which determines the groupings from $\mathcal{L}$.
Running \cref{alg:ordering} is rather efficient, depending on the number of factors only and needing to compute pairwise distances only once.

\begin{algorithm}[htb]
	\caption{Hierarchical Advanced Colour Passing}
	\label{alg:hierACP}
	\alginput{An \ac{fg} $M = (\boldsymbol R \cup \boldsymbol \Phi, \boldsymbol E)$, an index $i\in\{1,\ldots,m-1\}$,\\\hspace*{\algorithmicindent} and the outcome of \cref{alg:ordering} run on $M$.\\} 
	\algoutput{A lifted representation $M'$, encoded as a parametric \\\hspace*{\algorithmicindent} \ac{fg}, which is approximately equivalent to $M$.\\} 
	\begin{algorithmic}[1]
        \LeftComment{Phase I: Load groups of pairwise $\varepsilon$-equivalent factors for $\varepsilon_i$}
        \State Let $\mathcal{L}$ be the current list of candidate groups.
        \For{$k = m+i$ \textbf{to} $m+1$}
            \If{$k$ occurs in any group in $\mathcal{L}$} 
                \State Load global parent group $G_p(k) \in \mathcal{L}$
                \State Store group of $\varepsilon$-equivalent factors:
                \Statex \quad \quad ~$\boldsymbol G_{\Phi}(k) := \{\phi_j \mid j \in G_p(k),\ j < m\} \in \boldsymbol G$
                \State Update $\mathcal{L} \gets \mathcal{L} \setminus G_p(k)$
            \EndIf
        \EndFor
        \ForEach{factor $\phi_j \in \boldsymbol \Phi\setminus\left(\bigcup_{k=m+1}^{m+i}G_{\Phi}(k)\right)$}
            \State Store group $\boldsymbol G_{\Phi}(j):=\{\phi_j\}\in \boldsymbol G$
		\EndFor\hspace*{\algorithmicindent}
        
        \LeftComment{Phase II: Assign colours to factors and run ACP}\;
		\ForEach{group $\boldsymbol G_j \in \boldsymbol G$}
			\ForEach{factor $\phi_i \in \boldsymbol G_j$}
				\State $\phi_i.colour \gets j$\;
			\EndFor
		\EndFor
		\State $M' \gets$ Call ACP on $M$ and $\boldsymbol G$ using the assigned colours\;\label{line:eacp_call_acp}\hspace*{\algorithmicindent}
		\LeftComment{Phase III: Update potentials}\;
		\ForEach{parametric factor $g\in M'$}
            \State $\boldsymbol G_j \gets$ Collect all factors $\phi_i$ that were merged into $g$
                \State $\phi^*(\boldsymbol r) \gets \frac{1}{\vert\boldsymbol G_j\vert} \sum_{\phi_i \in \boldsymbol G_j} \phi_i(\boldsymbol r)$ for all assignments $\boldsymbol r$\; \label{line:eacp_mean_potential}
                \ForEach{factor $\phi_i \in \boldsymbol G_j$} \label{line:eacp_for_each_loop_update_potentials_factors}
				    \State $\phi_i \gets \phi^*$\; \label{line:eacp_update_potentials}
                \EndFor
		\EndFor
	\end{algorithmic}
\end{algorithm}

\subsection{Hierarchical Advanced Colour Passing Algorithm}\label{subsection:HACP}

\ac{hacp} provides a hyperparameter-free hierarchical approach to lifted model construction.
It uses the output of \cref{alg:ordering} to determine for a given level which groups get the same colour assigned, which is then the input to standard \ac{acp}, which runs independent of $\varepsilon$.
Specifically, \ac{hacp} proceeds in three phases, loading groups, running \ac{acp}, and updating potentials.
\cref{alg:hierACP} shows an overview, which is specified for a given hierarchy level $i$ for the sake of brevity but could be easily extended to build parametric models for all levels of the hierarchy.
It takes an \ac{fg}, an index $i$, and the output of \cref{alg:ordering} for the \ac{fg} as input.

Phase I provides a systematic procedure for forming the groups of factors in the input \ac{fg} given the nested list $\mathcal{L}$ of the output of \cref{alg:ordering}.
For instance, consider \cref{example:alltogether} and level $4$ with $\varepsilon_4$. 
The grouping induced at this level is:
\begin{align*}
    \boldsymbol G = & \{\boldsymbol G_1=\{\phi_1,\phi_2,\phi_3,\phi_4\},\boldsymbol G_2=\{\phi_5,\phi_6\},\\
                    &\phantom{\{}\boldsymbol G_3=\{\phi_7\},\boldsymbol G_4=\{\phi_8\},\boldsymbol G_5=\{\phi_9\},\boldsymbol G_6=\{\phi_{10}\} \}.
\end{align*}
Subsequently, Phase~II applies the standard \ac{acp} with colours assigned to each identified group, which returns a parametric \ac{fg} with one parametric factor for each group of factors deemed equivalent. 
In Phase~III, the potentials of each parametric factor are replaced by the arithmetic mean of the group of equivalent factors.
That is, for each group of potentials $\{\varphi_1, \dots, \varphi_k\}$, a new potential is computed as  $\varphi^* = \argmin_{\hat{\varphi}} \sum_{i=1}^{k} (\varphi_i - \hat{\varphi})^2$.
This minimises intra-group variance and yields an optimal compressed representation. 
The constructed factor $\boldsymbol\phi^* = (\varphi_1^*, \ldots, \varphi_n^*)$ is, by design, also pairwise $\varepsilon$-equivalent to all original factors from its generating group.

\section{Hierarchical Bounds: Compression vs. Accuracy}\label{sec:bounded}
A crucial property of the \ac{eacp} algorithm is its induced bound on the change in probabilistic queries based on its hyperparameter $\varepsilon$. \Ac{hacp} uses predefined groups (selected via \Cref{alg:hierACP}) and still allows choosing a compressed model of reduced complexity, while preserving the same asymptotic bounds as its predecessor, \ac{eacp}, to ensure consistent and reliable performance (accuracy).
The following subsections explore the properties of \ac{hacp}, examining how to control this information to apply the algorithm with specific accuracy.

\subsection{Asymptotic Properties}

A desirable property of the \ac{hacp} algorithm is that it preserves the same boundaries on the change in probabilistic queries as the \ac{eacp} algorithm.
After running \ac{eacp}, the deviation-wise worst-case scenario for an assignment is bounded.
Notably, \ac{hacp} relies on the mean values of the potentials of pairwise $\varepsilon$-equivalent factors.
To quantify the difference in probabilistic queries between the original \ac{fg} and the hierarchical processed \ac{fg} after applying the \ac{hacp} algorithm, we use the symmetric distance measure between two distributions $P_M$ and $P_{M'}$ introduced by \citet{Chan2005a}, which effectively bounds the maximal deviation of any assignment $\boldsymbol r$:
\begin{align}
	D_{CD}(P_M, P_{M'}) := \ln \max_{\boldsymbol r} \frac{P_{M'}(\boldsymbol r)}{P_M(\boldsymbol r)} - \ln \min_{\boldsymbol r} \frac{P_{M'}(\boldsymbol r)}{P_M(\boldsymbol r)}. \label{eq:eacp_distance_measure}
\end{align}
For \ac{eacp}, the following asymptotic bound has been proven~\citep{Luttermann2025a}:

\begin{theorem}[\citet{Luttermann2025a}]\label{th:lowerboundeacp_error_bound_given_eps}
	Let $M = (\boldsymbol R \cup \boldsymbol \Phi, \boldsymbol E)$ be an \ac{fg} and let $M'$ be the output of \ac{eacp} when run on $M$. With $P_M$ and $P_{M'}$ being the underlying full joint probability distributions encoded by $M$ and $M'$, respectively, and $m = \vert\boldsymbol \Phi\vert$, it holds that
	\begin{align}
		D_{CD}(P_M, P_{M'})   &\leq 
        \ln\left(\frac {\big( 1+\frac{m-1}{m}\varepsilon \big) \big( 1+\varepsilon \big)}{1+\frac{1}{m}\varepsilon}\right)^m\label{eq:sharp}\\
		&< \ln \big( 1 + \varepsilon \big)^{2m}< \ln \left(\frac{1+\varepsilon}{1-\varepsilon}\right)^m,
	\end{align}
where the bound given in \cref{eq:sharp} is optimal (sharp).
\end{theorem}

We next show that this bound also applies to the \ac{hacp} algorithm.
\begin{proposition}\label{corollary:transtohier}
\cref{th:lowerboundeacp_error_bound_given_eps} holds the same way for $M'$ being the output of the \ac{hacp} algorithm (\cref{alg:hierACP}).
\end{proposition}
\begin{proof}
The core components of the \ac{eacp} algorithm and its hierarchical counterpart \ac{hacp} (\cref{alg:hierACP}) are identical, aside from enforcing predefined group structures to guarantee a hierarchical structure.
Therefore, the proof can be conducted in the same manner as the original proof \cite[App. A]{Luttermann2025a}. The same proposed example can be used to hit the bound of \cref{eq:sharp}, showing its optimality.
\end{proof}

\subsection{Compression versus Accuracy}
We continue to give a finer analysis of theoretical properties entailed by \ac{hacp} regarding the trade-off between compression and accuracy.
All upcoming results hold for both \ac{eacp} and \ac{hacp}.
\begin{theorem}\label{th:bound_the_deviation}
The maximal absolute deviation between any initial probability $p = P_M(r \mid \boldsymbol e)$ of $r$ given $\boldsymbol e$ in model $M$ and the probability $p'=P_{M'}(r \mid \boldsymbol e)$ in the modified model $M'$ resulting from running \ac{hacp} (\cref{alg:hierACP}) or \ac{eacp} on $M$ can be bounded by
\begin{align*}
  p_{\max\Delta} :=\max_{\text{for any }r \mid \boldsymbol e}\vert p-p'\vert
  \leq \frac{\sqrt{e^d}-1}{\sqrt{e^d}+1} \text{ with } d=D_{CD}(P_M,P{_M'}).
\end{align*}
\end{theorem}
\begin{proof}[Proof Sketch] 
From \citet{Chan2005a}, we use
\begin{align}
	\frac{p e^{-d}}{p(e^{-d} -1) + 1} \leq p' = P_{M'}(r \mid \boldsymbol e) \leq \frac{p e^{d}}{p(e^{d} -1) + 1} \label{eq:eacp_error_bound_prob}
\end{align}
to define a upper bound function $f_{\max\Delta}$ for $p_{\max\Delta}$, which is symmetric for $p'=1/2$ (see \cref{fig:upperlower})
\begin{align*}
    f_{\max\Delta}(p):=& \max(f_{\text{upper}}(p),f_{\text{lower}}(p)) \text{ for } p\in[0,1]\\
  \text{with }  f_{\text{upper}}(p) :=&\frac{pe^{d}}{p(e^{d}-1)+1} - p 
=\frac{p (1-p) (e^{d}-1)}{p(e^{d}-1)+1}\\
\text{and } f_{\text{lower}}(p) :=&p -\frac{pe^{-d}}{p(e^{-d}-1)+1} = \frac{p (1-p) (1-e^{-d})}{p(e^{-d}-1)+1},
\end{align*}
and calculate its extrema by first- and second-order conditions.
\begin{figure}[tb]
\begin{center}
\includegraphics[width=0.9\linewidth]{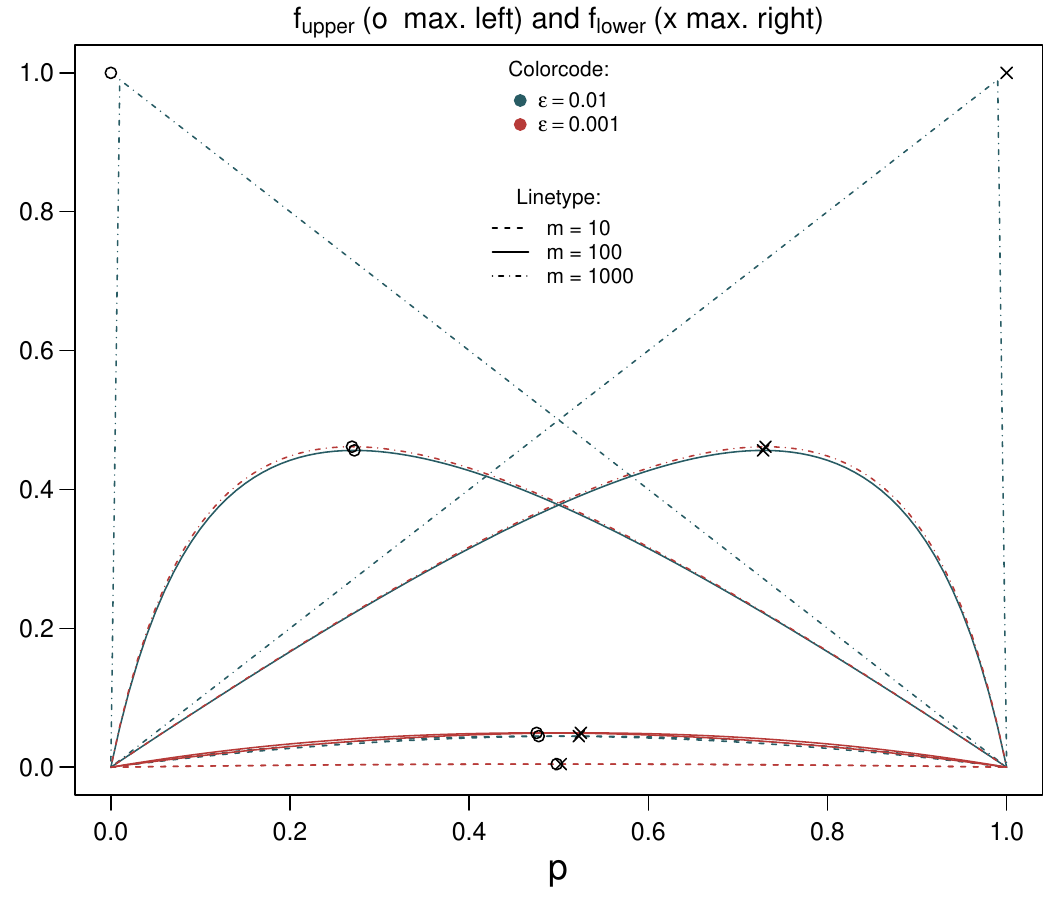}
\caption{Showing $f_{\text{upper}}$ and $f_{\text{lower}}$ over $[0,1]$ with $d_2$ values from \cref{corollary:inequalities} to use an upper estimate for $p_{\max\Delta}$ by bounding it from above.}\label{fig:upperlower}
\end{center}
\end{figure}
\end{proof}

Since \cref{th:bound_the_deviation}  lets us compute $D_{CD}(P_M,P_{M'})$ efficiently -— without inspecting the full factor graph -— it has direct practical impact.
Since $\frac{\sqrt{e^d}-1}{\sqrt{e^d}+1} = \tanh\left(\frac{d}{4}\right) \leq 1$ is monotonically strictly increasing in $d$, the results of \cref{th:lowerboundeacp_error_bound_given_eps} can be substituted into its formula, while the directions of the inequalities are obtained.
\begin{corollary}\label{corollary:inequalities}
With previous notations, the change in any probabilistic query in an initial model $M$ and a modified model $M'$ obtained by running \ac{hacp} (\cref{alg:hierACP}) or \ac{eacp} is bounded by
\begin{align*}
    p_{\max\Delta} &\leq \frac{\sqrt{e^{d_1}}-1}{\sqrt{e^{d_1}}+1} \text{ with } d_1=D_{CD}(P_M,P{_M'})\\
    &\leq \frac{\sqrt{e^{d_2}}-1}{\sqrt{e^{d_2}}+1}  \text{ with } d_2=\ln\left(\frac {\big( 1+\frac{m-1}{m}\varepsilon \big) \big( 1+\varepsilon \big)}{1+\frac{1}{m}\varepsilon}\right)^m\\
    &\leq \frac{\sqrt{e^{d_3}}-1}{\sqrt{e^{d_3}}+1} \text{ with } d_3= \ln \big( 1 + \varepsilon \big)^{2m}\\
    &\leq \frac{\sqrt{e^{d_4}}-1}{\sqrt{e^{d_4}}+1} \text{ with }     d_4= \ln \left(\frac{1+\varepsilon}{1-\varepsilon}\right)^m.
\end{align*}
\end{corollary}

\begin{figure}[tb] 
    \centering
    \includegraphics[width=.8\columnwidth]{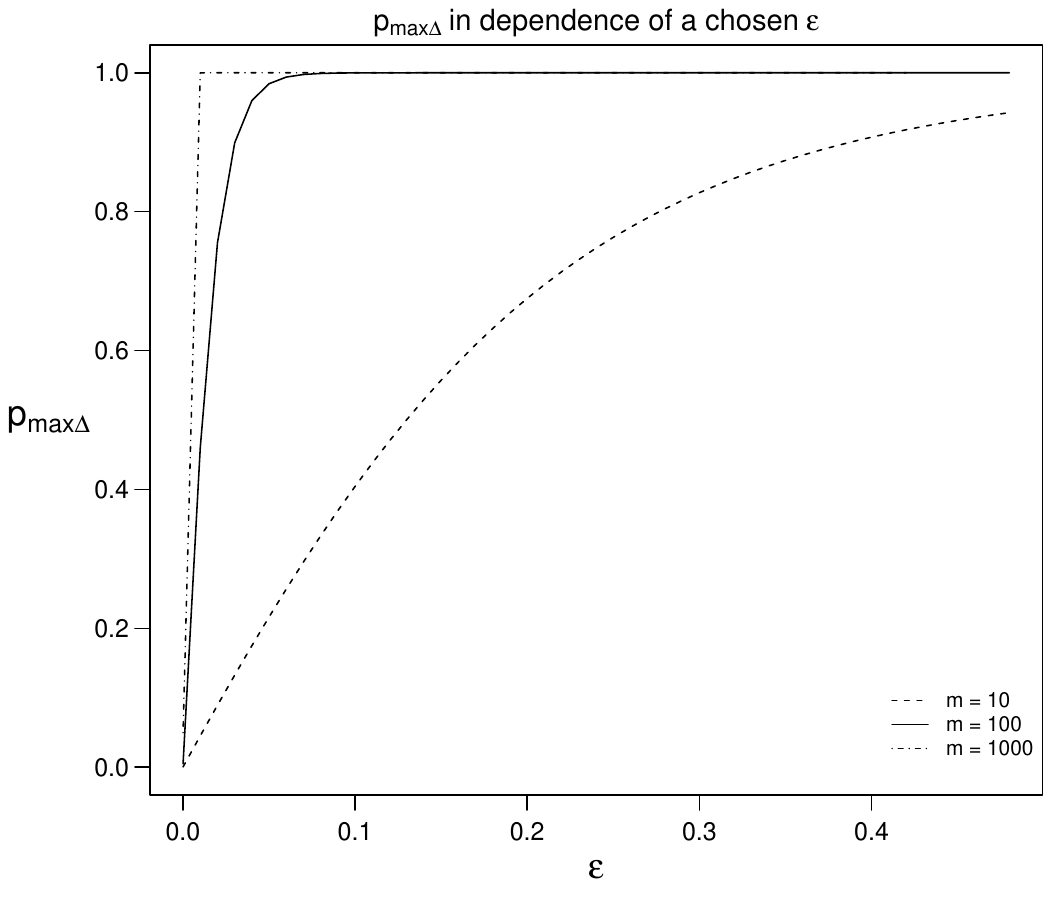}
    \caption{The bound on $p_{\max\Delta}$ depending on the choice of $\varepsilon$ for different amounts of factors $m$.}\label{fig:p_delta_given_epsilon}
\end{figure}

This implies that for a given $\varepsilon > 0$, we can determine the maximum deviation of $p_{\max\Delta}$ (see \cref{fig:p_delta_given_epsilon}).
However, it is also essential to consider the reverse perspective to gain insight into the overall diversity and structural complexity of the \ac{fg}.
Therefore, we investigate the following question:
What choice of $\varepsilon$ guarantees that our probabilities remain within a specified distance, i.e., $p_{\max\Delta} \leq p_{\Delta}^*$?
This question is addressed in the subsequent theorem by reversing the preceding inequalities and solving for $\varepsilon$.

\begin{theorem}\label{corollar:distancecorollar}
For any given $p_{\Delta}^*\in(0,\frac{1}{2}]$, the output of \ac{hacp} guarantees for any $\varepsilon \in (0,1)$, which is smaller or equal to
{\scriptsize
\begin{align*}
  \hspace{-0.05cm}\varepsilon_1 =-\frac{1+\tfrac{m-1}{m}-\frac{1}{m}\sqrt[m]{e^d}}{2\,\frac{m-1}{m}}
  +\sqrt{\Bigl(-\frac{1+\tfrac{m-1}{m}-\frac{1}{m}\sqrt[m]{e^d}}{2\,\frac{m-1}{m}}\Bigr)^2
    -\frac{1-\sqrt[m]{e^d}}{\tfrac{m-1}{m}}}
\end{align*}
}
with $d =\ln \left(\frac{p_{\Delta}^*+1}{1-p_{\Delta}^*}\right)^2$ 
 the bound $p_{\max\Delta}\leq p_{\Delta}^*$. 
\end{theorem}

\begin{proof}
Using \cref{th:bound_the_deviation}, we get for $d=D_{CD}(P_M,P{_M'})$:
\begin{align*}
&p_{\max\Delta} \leq \frac{\sqrt{e^d}-1}{\sqrt{e^d}+1}=:p_{\Delta}^* \\
&\Leftrightarrow   \ln \left(\frac{p_{\Delta}^*  +1}{1- p_{\Delta}^* } \right)^2 = d=D_{CD}(P_M,P{_M'}).
\end{align*}

 Additionally, we use \cref{eq:sharp} from \cref{th:lowerboundeacp_error_bound_given_eps} and the corresponding version for \ac{hacp} (\cref{corollary:transtohier}) and solve the inequality for $\varepsilon$, when it reaches equality:
\begin{align*}
    \quad\quad\quad\quad\quad~~~ d = \ln\left(\frac {\big( 1+\frac{m-1}{m}\varepsilon \big) \big( 1+\varepsilon \big)}{1+\frac{1}{m}\varepsilon}\right)^m.~~~~\quad\quad\quad\quad\raisebox{-2.9ex}{\qedhere}
    \end{align*}
\end{proof}    
This means that we can bound the maximal deviation $p_{\Delta}^*$, which is tight \cite{Luttermann2025a}, of \ac{hacp} and \ac{eacp}, respectively, by calculating $\varepsilon_1(p_{\Delta}^*)$ before we run it. 
However, these bounds are mostly of theoretical nature and the deviations are rarely encountered in reality \cite{Luttermann2025a}. 
\begin{note}
Choosing, for instance, a $10$ times larger $\varepsilon$ has pretty much the same effect as choosing $10$ times the number of factors $m$ on the bound on the change in probabilistic queries (cf. \cref{fig:upperlower}).
\end{note} 

Thus, this subsection lays the groundwork for leveraging the supplementary insights from \cref{alg:ordering} to develop a holistic understanding of the structural and computational complexity of the \ac{fg}.
Depending on sensitivity to $\varepsilon$, $p_{\max\Delta}/p_{\Delta}^*$, and $m$, one can assess the implications of specific parameter choices on the effect of compressing a given \ac{fg}, or alternatively, prioritise model fidelity.
The \ac{hacp} algorithm enables such assessments across multiple structural levels.

\section{Discussion}
\paragraph{Related work.}
Lifted inference exploits the indistinguishability of objects in 
probabilistic (relational) models, enabling more efficient query answering (marginals of \acp{rv} given observations) with exact results~\citep{NieBr14}.
First introduced by \citet{Poo03}, parametric \acp{fg}, which combine relational logic and probabilistic modelling, and lifted variable elimination enable lifted probabilistic inference to speed up query answering by exploiting the indistinguishability of objects.
Over the past years, lifted variable elimination has continuously been refined by many researchers to reach its current form~\citep{BraMo18b,DeSalvoBraz2005a,DeSalvoBraz2006a,Kisynski2009a,Milch2008a,TagFiDaBl13}.
To construct a lifted (i.e., first-order) representation such as a parametric \ac{fg}, the \ac{acp} algorithm~\citep{LutBrMoGe24}, which generalises the CompressFactorGraph algorithm~\citep{AhmKeMlNa13,Kersting2009a}, is the current state of the art.
\Ac{acp} runs a colour passing procedure to detect symmetric subgraphs in a probabilistic graphical model, similar to the Weisfeiler-Leman algorithm~\citep{Weisfeiler1968a}, which is a well-known algorithm to test for graph isomorphism.
It groups symmetric subgraphs and exploits exponentiation during probabilistic inference.
While \ac{acp} is able to construct a parametric \ac{fg} entailing equivalent semantics as a given propositional model, it requires potentials of factors to exactly match before grouping them.
In practice, however, potentials are often estimates and hence might slightly differ even for indistinguishable objects. 
To account for small deviations between potentials, the \ac{eacp} algorithm~\citep{Luttermann2025a} has been introduced, generalising \ac{acp} by introducing a hyperparameter $\varepsilon$ that controls the trade-off between the exactness and the compactness of the resulting lifted representation.
While the original formulation assumes factors with identical dimensions and range structures, the general concept naturally extends to heterogeneous dimensions.
In such cases, care must be taken to preserve structural consistency to ensure that existing symmetries remain valid when reasoning across dimensions.

\paragraph{Pre‐ordering means pre‐analysing.}
Our hierarchical algorithm (\cref{alg:ordering}) imposes a predetermined nesting structure on the factor graph before any colour passing procedure, enabling a priori application-specific level selection.
By 
specifying levels before applying \cref{alg:hierACP} to the adjusted graph, one can predict and implicitly control the resulting complexity of $\varepsilon$-equivalent group structure and thereby enhance interpretability.
In contrast to \ac{eacp}, which may produce $\varepsilon$-equivalent groupings that lack consistent nesting across runs or parameter settings, our hierarchical approach (\ac{hacp}) ensures structural coherence and comparability across instances.
Moreover, the explicit composition of each level can be monitored to trace modification impacts throughout the hierarchy.
Large $d_\infty$ values indicate low symmetry and are generally unsuitable for approximating the \ac{fg} in most applications. Conversely, many small $d_\infty$ values close to each other suggest high potential for similar factor structures.
Such an analysis is not feasible with \ac{eacp}, which requires an a priori choice of $\varepsilon$ without any guarantees of identifying symmetries. 

\paragraph{Trade‐off: Compression versus Accuracy.}
\ac{eacp} and \ac{hacp} inherit deviation bounds from \citet{Luttermann2025a}, yielding identical sharp bounds and dependencies for any choice of $\varepsilon$.
In practice, grouping composition controls magnitude and sign of probabilistic deviations in downstream queries:
As the hierarchy level (or $\varepsilon$) increases, theoretical bounds grow, yet actual query deviations may fluctuate based on group aggregations.
Crucially, our hierarchical bounds facilitate pre‐specification of maximal permissible $\varepsilon$ values and corresponding levels.
Rather than relying on the generally intractable $D_{CD}$ for approximated models \cite{Chan2005a}, one can derive $p_{\max\Delta}$ for a given $\varepsilon$, or select an admissible level that guarantees both desired compression and sufficient accuracy.

\ac{hacp} operates on a more restricted space of $\varepsilon$-equivalent groupings than \ac{eacp} due to 
its forced hierarchy, 
enabling interpretability and structured level comparisons. 
Thus, \ac{hacp} may show slightly higher average deviations without systematic inferiority on individual assignments.
Importantly, both algorithms retain identical worst-case deviation bounds by construction.

\section{Conclusion}
We introduce a novel framework for hierarchical lifting and model reconciliation in \acp{fg}.
By presenting a more practical one-dimensional notion of $\varepsilon$-equivalent factors, we enable the identification of (possibly inexact) symmetries, the number and sizes of $\varepsilon$-equivalent groups and the resulting reduction of computational complexity, thereby allowing for lifted inference.
Our theoretical analysis provides a solid foundation for understanding the structural properties of \acp{fg}.
Crucially, the entire hierarchy is fixed prior to initiating colour passing or inference, ensuring structural consistency and enabling theoretical error bounds.
This work provides a foundation for future advances in efficient and interpretable probabilistic inference.

\begin{ack}
This work was partially funded by the Ministry of Culture and Science of the German State of North Rhine-Westphalia.
The research of Malte Luttermann was funded by the BMBF project AnoMed 16KISA057.
\end{ack}

\section*{Appendix}
\addcontentsline{toc}{section}{Appendix}
\setcounter{section}{0}
\section{Counterexamples}
\begin{proposition}
    The \ac{odeed} is not a metric.
\end{proposition}
\begin{proof}
    The \ac{odeed} 
    is not a metric, because the $\Delta$-inequality does not hold, which is exemplarily proven by this counterexample:
    \begin{align*}
        \phi_1 &= \begin{pmatrix} 2 \\ 0.5 \end{pmatrix},\phi_2 = \begin{pmatrix} 1 \\ 1 \end{pmatrix},  \phi_3 = \begin{pmatrix}  1 \\ 2 \end{pmatrix} 
    \end{align*}
    with $d_{\infty}(\phi_1,\phi_2)+d_{\infty}(\phi_2,\phi_3) = 1+1 < 3= d_{\infty}(\phi_1,\phi_3)$.
\end{proof}

\begin{proposition}
    The \ac{odeed} lacks the transitivity property.
\end{proposition}
\begin{proof}
    Given a Boolean random variable $Var$, consider three factors $\phi_1,\phi_2,\phi_3$ defined as follows:
    \begin{center}
        \begin{tabular}{cccc}
            \toprule
                $Var$        & $\phi_1(Var)$ & $\phi_2(Var)$ & $\phi_3(Var)$ \\ 
            \midrule
                \text{true}  & $0.95$        & $1.0$          & $1.08$ \\
                \text{false} & $2.05$        & $1.95$         & $2.10$ \\ 
            \bottomrule
        \end{tabular}
    \end{center}
    Using the \ac{odeed} with $\varepsilon=0.1$, we end up with $\phi_1=_{\varepsilon}\phi_2$ and $\phi_2=_{\varepsilon}\phi_3$, but $\phi_1\neq_{\varepsilon}\phi_3$:
    \begin{align*}
        \phi_1(\text{true})=0.95< 0.972=(1-\varepsilon) 1.08 =(1-\varepsilon) \phi_3(\text{true}).
    \end{align*}
\end{proof}

\section{Detailed Proofs}
\setcounter{theorem}{1}
\begin{theorem}
Two vectors $\phi_1,\phi_2\in\mathbb{R}_{>0}^n$ are $\varepsilon$-equivalent (\Cref{def:original_epsilon_equivalence}) if and only if $d_{\infty}(\phi_1,\phi_2)\leq \varepsilon$ holds.
\end{theorem}
\begin{proof}
    In mathematical terms, the claim can be summarised as $\phi_1 =_{\varepsilon}\phi_2 \Leftrightarrow d_{\infty}(\phi_1,\phi_2)\leq \varepsilon$, which we prove for any $\varepsilon>0$:
    \begin{align*}
        &\phi_1 =_{\varepsilon}\phi_2 \text{ for two factors } \phi_1,\phi_2\in\mathbb{R}_{>0}^n\\
        &\overset{\text{def.}}{\Leftrightarrow} \phi_1(k)\in[(1-\varepsilon)\phi_2(k), (1+\varepsilon)\phi_2(k)] \text{ and }\\
        & \quad ~~~\phi_2(k)\in[(1-\varepsilon)\phi_1(k), (1+\varepsilon)\phi_1(k)] \text{ for }k=1,\ldots,n\\
        &\Leftrightarrow \phi_2(k)-\phi_2(k)\varepsilon \leq \phi_1(k)\leq \phi_2(k)+\varepsilon\phi_2(k) \text{ and }\\
        &\quad ~~~\phi_1(k)-\phi_1(k)\varepsilon \leq \phi_2(k)\leq \phi_1(k)+\varepsilon\phi_1(k)\text{ for }k=1,\ldots,n\\
        &\Leftrightarrow -\phi_2(k)\varepsilon\leq\phi_1(k)-\phi_2(k)\leq\varepsilon\phi_2(k)\text{ and }\\
        &\quad ~~~-\phi_1(k)\varepsilon\leq\phi_2(k)-\phi_1(k)\leq\varepsilon\phi_1(k)\text{ for }k=1,\ldots,n\\
        &\Leftrightarrow \vert \phi_1(k)-\phi_2(k)\vert \leq \varepsilon \phi_2(k)\text{ and } \\
        &\quad ~~~ \vert \phi_1(k)-\phi_2(k)\vert \leq \varepsilon \phi_1(k)\text{ for }k=1,\ldots,n\\
        &\Leftrightarrow \frac{\vert \phi_1(k)-\phi_2(k)\vert}{ \phi_2(k)} \leq \varepsilon\text{ and }\\
        &\quad ~~~\frac{\vert \phi_1(k)-\phi_2(k)\vert}{\phi_1(k)} \leq \varepsilon \text{ for }k=1,\ldots,n\\
        &\Leftrightarrow\frac{\vert \phi_1(k)-\phi_2(k)\vert}{ \min\{\phi_1(k),\phi_2(k)\} }  \leq \varepsilon\text{ for }k=1,\ldots,n\\
        &\Leftrightarrow\max_{k=1,\ldots,n}\left\{\frac{\vert \phi_1(k)-\phi_2(k)\vert}{ \min\{\phi_1(k),\phi_2(k)\} }\right\}  \leq \varepsilon\\
        &\overset{\text{def.}}{\Leftrightarrow} d_{\infty}(\phi_1,\phi_2) \leq \varepsilon\qedhere
    \end{align*}
\end{proof}

\setcounter{theorem}{4}
\begin{theorem}\label{th:bound_the_deviation222}
The maximal absolute deviation between any initial probability $p = P_M(r \mid \boldsymbol e)$ of $r$ given $\boldsymbol e$ in model $M$ and the probability $p'=P_{M'}(r \mid \boldsymbol e)$ in the modified model $M'$ resulting from running Hierarchical Advanced Colour Passing (HACP, main paper, Alg. 2) or $\varepsilon$-Advanced Colour Passing ($\varepsilon$-ACP) on $M$ can be bounded by
\begin{align*}
  p_{\max\Delta} :=\max_{\text{for any }r \mid \boldsymbol e}\vert p-p'\vert
  \leq \frac{\sqrt{e^d}-1}{\sqrt{e^d}+1} \text{ with } d=D_{CD}(P_M,P{_M'}).
\end{align*}
\end{theorem}
\begin{proof} 
From Chan and Darwiche~\cite{appChan2005a}, we already know that
\begin{align}
	\frac{p e^{-d}}{p(e^{-d} -1) + 1} \leq p' = P_{M'}(r \mid \boldsymbol e) \leq \frac{p e^{d}}{p(e^{d} -1) + 1} \label{eq:eacp_error_bound_prob222}
\end{align}
holds (see \cref{fig:upperandlowerboundp222}), where $p = P_M(r \mid \boldsymbol e)$ is the probability of $r$ given $\boldsymbol e$ in the original model $M$ and $d = D_{CD}(P_M,P{_M'})$ is the value of the distance measure introduced by Chan and Darwiche between $P_M$ and $P{_M'}$. Hence, for any $r$ given $\boldsymbol e$, in the worst case, we get
\begin{align*}
   \vert p-p'\vert = \begin{cases}
       p'-p & \text{for } p \leq  p'\\
       p-p' & \text{for } p'< p
   \end{cases}
   =\begin{cases}
       \frac{pe^{d}}{p(e^{d}-1)+1} - p & \text{for } p \leq p'\\
       p-\frac{pe^{-d}}{p(e^{-d}-1)+1}& \text{for } p'< p.
   \end{cases}
\end{align*}
\begin{figure}[t]
\begin{center}
\includegraphics[width=\linewidth]{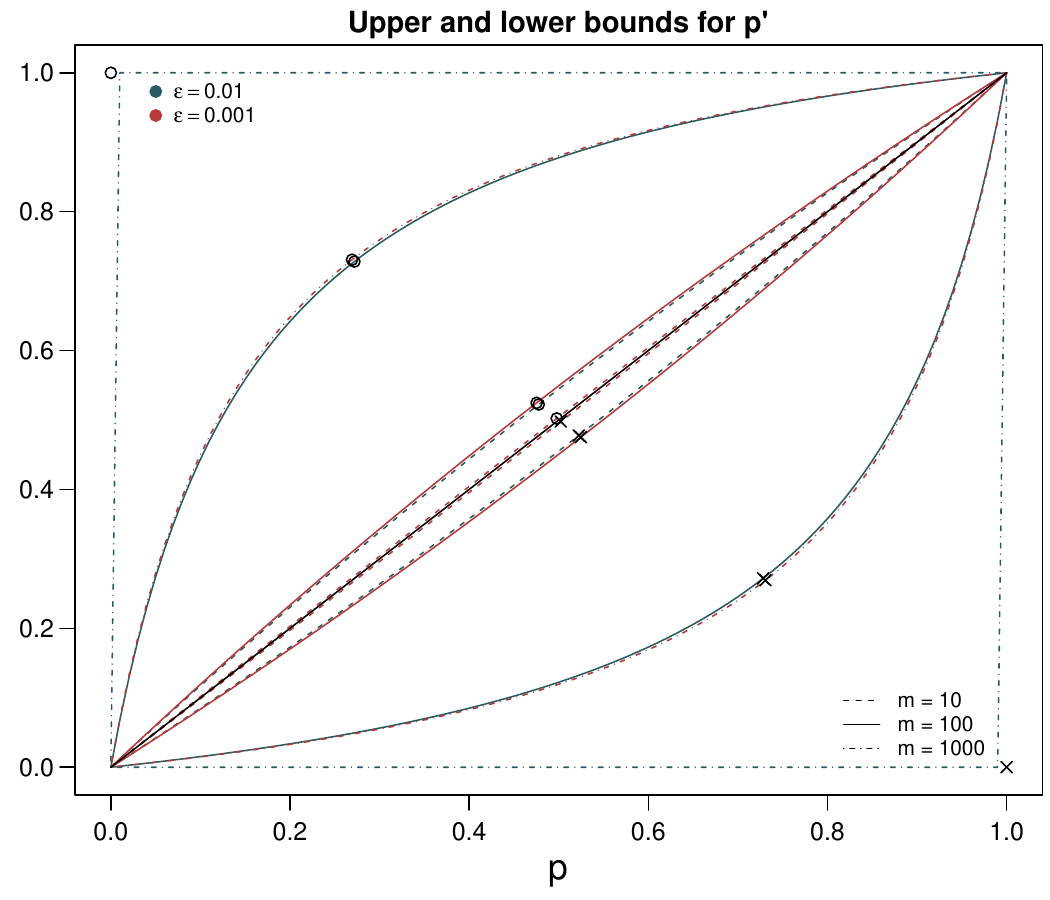}\vspace{-0.5cm} 
\caption{Bounds of \cref{eq:eacp_error_bound_prob222} in comparison to $p$ for different $d$ values depending on $m$ and $\varepsilon$ (Eq. (5) of the main paper). 
Circles/ crosses are the maximum distances from those bounds to the function $p$. Distances to function $f(p)=p$ are later also referred as $f=f_{\text{upper}}$ for $p\in[0,\frac{1}{2}]$ and $f=f_{\text{lower}}$ for $p\in(\frac{1}{2},1]$.}\label{fig:upperandlowerboundp222}
\end{center}
\end{figure}
Becoming independent of $p'$ guarantees one maximal bound for all possible queries and can be achieved using the maximum of both cases in $[0,1]$ as an upper bound for $ p_{\max\Delta}$, which is given by the following function $f_{\max\Delta}(p)$: 
\begin{align*}
    f_{\max\Delta}(p):=& \max(f_{\text{upper}}(p),f_{\text{lower}}(p)) \text{ for } p\in[0,1]\\\\
  \text{with }  f_{\text{upper}}(p) :=&\frac{pe^{d}}{p(e^{d}-1)+1} - p 
=\frac{p (1-p) (e^{d}-1)}{p(e^{d}-1)+1}\\
\text{and } f_{\text{lower}}(p) :=&p -\frac{pe^{-d}}{p(e^{-d}-1)+1} = \frac{p (1-p) (1-e^{-d})}{p(e^{-d}-1)+1}.
\end{align*}
\begin{figure}[t]
\begin{center}
\includegraphics[width=\linewidth]{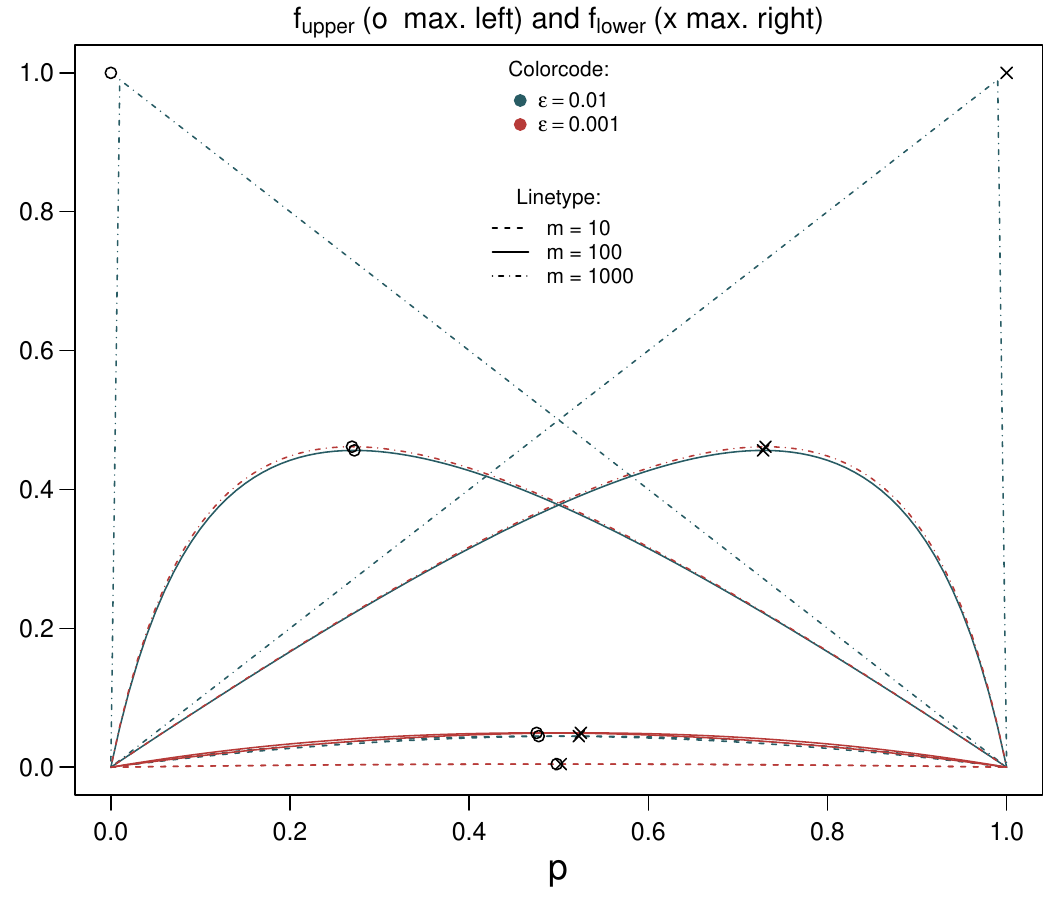}\vspace{-0.5cm} 
\caption{Showing the $f_{\text{upper}}$ and $f_{\text{lower}}$ functions over $[0,1]$ with $d_2$ values from Corollary~6 to use an upper estimate for $p_{\max\Delta}$ by bounding it from above.}\label{fig:upperlower222}
\end{center}
\end{figure}
It is easy to see that $f_{\max\Delta}$ is a symmetric function around $p=0.5$ (see \cref{fig:upperlower222}), because $f_{\text{upper}}(p)= f_{\text{lower}}(1-p)$ holds:
\begin{align*}
 f_{\text{lower}}(1-p) &= \frac{(1-p) p (1-e^{-d})}{(1-p)(e^{-d}-1)+1}\\
&= \frac{ (1-p) p (e^{d}-1)e^{-d}}{(1-p)(e^{-d}-1)+1}\\
&= \frac{ (1-p) p (e^{d}-1)}{((1-p)(e^{-d}-1)+1 )e^{d}}\\
&= \frac{ p (1-p) (e^{d}-1)}{(1-p)(1-e^{d})+e^{d}}\\
&=  \frac{ p (1-p) (e^{d}-1)}{1-p+pe^d}\\
&= \frac{ p (1-p) (e^{d}-1)}{p(e^d-1)+ 1}\\
&=f_{\text{upper}}(p)
\end{align*}
Now, the choice for $p'$ to get the maximum of both functions is $p=0.5$, while $f_{\text{upper}}(p)$ decreases for $p>0.5$ and $f_{\text{lower}}(p)$ increases for $ p<0.5$ for $d\geq 0$. 
Therefore, we get
\begin{align}
f_{\max\Delta}(p):= \begin{cases}
    f_{\text{upper}}(p) & \text{for } 0\leq p\leq 0.5\\
    f_{\text{lower}}(p) & \text{for } 0.5 <p\leq 1\,\ 
    \end{cases}
\end{align}
This means that our search for the maximum deviation leads us to calculate the derivatives after $p$
\begin{align*}
f_{\text{upper}}'(p) &= -\frac{(e^d-1) (p^2(e^d-1)+2p-1)}{(p(e^{-d}-1)+1)^2},\\
f_{\text{upper}}''(p) &= \frac{-2e^d(e^d-1)}{(p(e^{d}-1)+1)^3},\\
f_{\text{lower}}'(p) &= \frac{(1-e^{-d})(p^2(1-e^{-d})-2p+1)}{(p(e^{-d}-1)+1)^2},\\
f_{\text{lower}}''(p) &= \frac{-2e^d(e^d-1)}{(p+e^d(1-p))^3}, 
\end{align*}
and to consider initially the first-order conditions. For this purpose, we first obtain\vspace{-0.5cm}
\begin{align*}
f_{\text{upper}}'(p) &=0\\
	\Leftrightarrow &p^2(e^d-1)+2p-1  = 0 \\
    \Leftrightarrow &p_{\text{upper }1/2} = -\frac{1}{e^d-1}\pm \sqrt{\left(\frac{-1}{e^d-1}\right)^2+\frac{1}{e^d -1}}\\
    \Leftrightarrow &p_{\text{upper }1/2}= -\frac{1}{e^d-1}\pm \frac{\sqrt{e^d}}{e^d-1} = \frac{-1\pm\sqrt{e^d}}{e^d-1}.\\
 \end{align*}
 As $p_{\text{upper }2}$ is smaller than zero, the potential maximum in $[0,1]$ is at
 \begin{align*}
     p_1=p_{\text{upper }1}=\frac{\sqrt{e^d}-1}{e^d-1}=\frac{1}{\sqrt{e^d}+1}.
 \end{align*}
Analogously, we find a potential maximum for $f_{\text{lower}}'$:
    \begin{align*}
    f_{\text{lower}}'(p) &= 0\\
    \Leftrightarrow &p^2(1-e^{-d})-2p+1 =0\\
    \Leftrightarrow &p_{\text{lower }1/2} = \frac{1}{1-e^{-d}} \pm \sqrt{\left(\frac{1}{1-e^{-d}}\right)^2-\frac{1}{1-e^{-d}}}\\
    \Leftrightarrow &p_{\text{lower }1/2} = \frac{1}{1-e^{-d}}\pm \frac{\sqrt{e^{-d}}}{1-e^{-d}} =\frac{1\pm\sqrt{e^{-d}}}{1-e^{-d}}
\end{align*}
As $p_{\text{lower }1}$ is larger than one, the possible maximum in $[0,1]$ is at
\begin{align*}
    p_2= p_{\text{lower }2} = \frac{1-\sqrt{e^{-d}}}{1-e^{-d}} = \frac{1}{\sqrt{e^{-d}}+1} =\frac{\sqrt{e^{d}}}{\sqrt{e^{d}}+1}
\end{align*}
and the second-order conditions can also be easily checked:\\
Since $e^{d}-1>0$ and $1-p>0$, we get $f_{\text{upper}}''(p_1)<0$ and $f_{\text{lower}}''(p_2)<0$ and can conclude that $p_1$ is a local maximum of $f_{\text{upper}}$ and $p_2$ is a local maximum of $f_{\text{lower}}$. The boundary values $0$ and $1$ are no possible points for a global maximum, because both functions $f_{\text{upper}}$ and $f_{\text{lower}}$ take on the value $0$ there. Therefore, the only possible extreme point for the global maximum for $f_{\text{upper}}$ is $p_{1}=\frac{1}{\sqrt{e^d}+1}$ and $f_{\text{lower}}$ is $p_{2}=\frac{\sqrt{e^d}}{\sqrt{e^d}+1}$. Note that $p_1$ and $p_2$ are symmetrically distanced to $p = 1/2$.\\
Both reach exactly the same maximal deviation:
\begin{align*}
    f_{\text{upper}}(p_1) &=  \frac{\frac{1}{\sqrt{e^d}+1} (1-\frac{1}{\sqrt{e^d}+1}) (e^{d}-1)}{\frac{1}{\sqrt{e^d}+1}(e^{d}-1)+1} \\
    &=\frac{\frac{1}{\sqrt{e^d}+1}\cdot \frac{\sqrt{e^d}}{\sqrt{e^d}+1}\cdot (\sqrt{e^d}+1)\cdot (\sqrt{e^d}-1)}{\frac{e^d-1+\sqrt{e^d}+1}{\sqrt{e^d}+1}}\\
    &=\frac{\sqrt{e^d}(\sqrt{e^d}-1)}{e^d+\sqrt{e^d}} =\frac{\sqrt{e^d}-1}{\sqrt{e^d}+1}\\
    &\text{\hspace{-1.5cm}and the same holds for}\\
    f_{\text{lower}}(p_2) &= \frac{ \frac{\sqrt{e^d}}{\sqrt{e^d}+1} (1- \frac{\sqrt{e^d}}{\sqrt{e^d}+1}) (1-e^{-d})}{ \frac{\sqrt{e^d}}{\sqrt{e^d}+1}(e^{-d}-1)+1}\\
    &=\frac{\frac{\sqrt{e^d}}{\sqrt{e^d}+1}\cdot \frac{1}{\sqrt{e^d}+1}\cdot (1-e^{-d})}{\frac{\sqrt{e^d}}{\sqrt{e^d}+1} \cdot (\sqrt{e^{-d}}-\sqrt{e^d}+\sqrt{e^d}+1)}\\
     &= \frac{\sqrt{e^d} (1-\sqrt{e^{-d}})(1+\sqrt{e^{-d}})}{(\sqrt{e^{-d}}+1)(\sqrt{e^d}+1)}\\
      &= \frac{\sqrt{e^d}-1}{\sqrt{e^d}+1}.
\end{align*}
This means:
\begin{align*}
p_{\max\Delta} &=\max_{\text{for any }r \mid \boldsymbol e}\vert p-p'\vert \\&\leq f_{\text{upper}}(p_1) = f_{\text{lower}}(p_2) \\
&= \frac{\sqrt{e^d}-1}{\sqrt{e^d}+1} \qedhere
\end{align*}
\end{proof}

\setcounter{theorem}{6}
\begin{theorem}\label{corollar:distancecorollar2222}
For any given $p_{\Delta}^*\in(0,\frac{1}{2}]$, the output of HACP guarantees for any $\varepsilon \in (0,1)$, which is smaller or equal to
\begin{align*}
    \varepsilon_1 = & -\frac{1+\frac{m-1}{m}-\frac{1}{m}\sqrt[m]{e^d}}{2\frac{m-1}{m}} \\
    &+\sqrt{\left(-\frac{1+\frac{m-1}{m}-\frac{1}{m}\sqrt[m]{e^d}}{2\frac{m-1}{m}}\right)^2-\frac{1-\sqrt[m]{e^d}}{\frac{m-1}{m}}}
\end{align*}
with 
\begin{align*}
    d =\ln \left(\frac{p_{\Delta}^*+1}{1-p_{\Delta}^*}\right)^2
\end{align*}
 the bound $p_{\max\Delta}\leq p_{\Delta}^*$. 
\end{theorem}
This means that we can bound the maximal deviation $p_{\Delta}^*$ of HACP and $\varepsilon$-ACP, respectively, by calculating $\varepsilon_1(p_{\Delta}^*)$ before we run it. In \cite{appLuttermann2025a}, it is shown that the bound is tight.
\begin{proof}
Using \cref{th:bound_the_deviation222}, we get for $d=D_{CD}(P_M,P{_M'})$:
\begin{align*}
&p_{\max\Delta} \leq \frac{\sqrt{e^d}-1}{\sqrt{e^d}+1}=:p_{\Delta}^* \\
&\Leftrightarrow p_{\Delta}^* \left(\sqrt{e^d}+1\right)=\sqrt{e^d}-1\\
&\Leftrightarrow  p_{\Delta}^*  + 1 = (1- p_{\Delta}^* )\sqrt{e^d} \\ 
&\Leftrightarrow   \frac{p_{\Delta}^*  +1}{1- p_{\Delta}^* } = \sqrt{e^d}\\
&\Leftrightarrow   \ln \left(\frac{p_{\Delta}^*  +1}{1- p_{\Delta}^* } \right)^2 = d=D_{CD}(P_M,P{_M'}).
\end{align*}
Additionally, we know from Theorem~3 and the corresponding version for HACP (Proposition~4) that 
\begin{align} 
    D_{CD}(P_M,P{_M'})\leq \ln\left(\frac {\big( 1+\frac{m-1}{m}\varepsilon \big) \big( 1+\varepsilon \big)}{1+\frac{1}{m}\varepsilon}\right)^m.
\end{align}
Thus, the question we now answer is for which $\varepsilon$ this inequality reaches equality:
\begin{align*}
    & d = \ln\left(\frac {\big( 1+\frac{m-1}{m}\varepsilon \big) \big( 1+\varepsilon \big)}{1+\frac{1}{m}\varepsilon}\right)^m\\
    \Leftrightarrow &~ \sqrt[m]{e^d}= \frac {\big( 1+\frac{m-1}{m}\varepsilon \big) \big( 1+\varepsilon \big)}{1+\frac{1}{m}\varepsilon}\\
     \Leftrightarrow &~ \big(1+\frac{1}{m}\varepsilon\big) \sqrt[m]{e^d} = \big( 1+\frac{m-1}{m}\varepsilon \big) \big( 1+\varepsilon \big)\\
     \Leftrightarrow &~ 0= \frac{m-1}{m}\varepsilon^2 +\big(1+\frac{m-1}{m}-\frac{1}{m}\sqrt[m]{e^d} \big)\varepsilon +1-\sqrt[m]{e^d},
\end{align*}
which can be solved for $\varepsilon$ with $q_1=\frac{1+\frac{m-1}{m}-\frac{1}{m}\sqrt[m]{e^d} }{\frac{m-1}{m}}$ and $q_2=\frac{1-\sqrt[m]{e^d} }{\frac{m-1}{m}}$, resulting in
\begin{align*}
    \varepsilon_{1/2} = -\frac{q_1}{2}\pm \sqrt{\big(\frac{q_1}{2}\big)^2-q_2}.
\end{align*}
Since $q_1\geq 0 \Leftrightarrow \frac{m-1}{m}\geq p^*_{\Delta}$, the minus option $\varepsilon_2$ is smaller than $0$ and knowing that $m\geq 2$ already guarantees the result of \cref{corollar:distancecorollar2222}, for all cases which make sense to apply (better than guessing $\geq 0.5$), the only reasonable solution is $\varepsilon_1$.
\end{proof}

\begin{figure}[tb]
\begin{center}
\includegraphics[width=\linewidth]{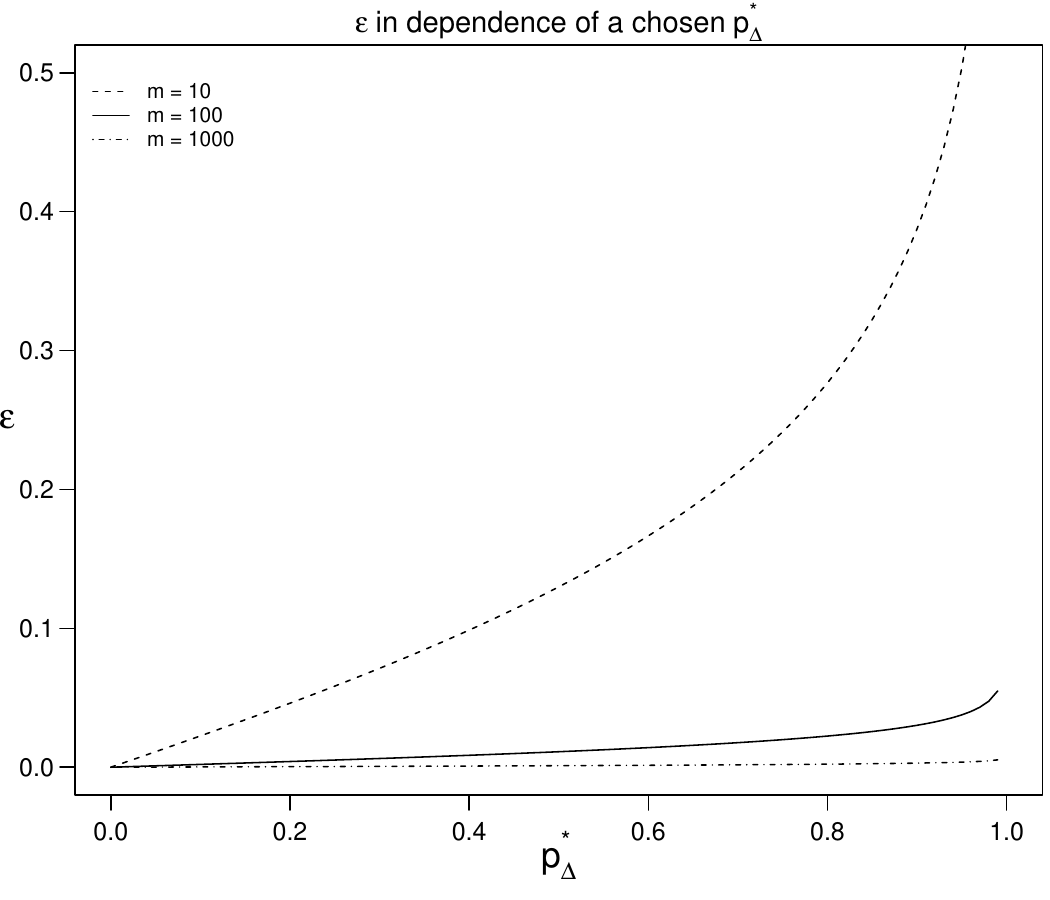}\vspace{-0.5cm} 
\caption{The maximal choice of  $\varepsilon$ depending on the maximal deviation $p_{\Delta}^*$ for different amount of factors $m$ to guarantee $p_{\max\Delta}\leq p_{\Delta}^*$ as proven in \cref{corollar:distancecorollar2222}.}\label{fig:p_delta_epsilon222}
\end{center}
\end{figure}
\section{The Basic Idea of Lifting} \label{app:lifting_example222}
To illustrate the idea behind lifting, consider the following example.

\begin{example}
	Take a look at the \ac{fg} illustrated in \cref{fig:example_fg} of the main paper and assume we want to answer the query $P(B = \text{true})$.
    We obtain
	\begin{align*}
		P(B = \text{true})
		&= \sum_{a \in \text{range}(A)} \sum_{c \in \text{range}(C)} P(A = a, B = \text{true}, C = c) \\
		&= \frac{1}{Z} \sum_{a \in \text{range}(A)} \sum_{c \in \text{range}(C)} \phi_1(a, \text{true}) \cdot \phi_2(c, \text{true}) \\
		&= \frac{1}{Z} \Big( \varphi_1 \varphi_1 + \varphi_1 \varphi_3 + \varphi_3 \varphi_1 + \varphi_3 \varphi_3 \Big).
	\end{align*}
	Since $\phi_1(A,B)$ and $\phi_2(C,B)$ are equivalent (in particular, it holds that $\phi_1(a, \text{true}) = \phi_2(c, \text{true})$ for all assignments where $a = c$), we can exploit this property to simplify the computation and get
	\begin{align*}
		P(B = \text{true})
		&= \frac{1}{Z} \sum_{a \in \text{range}(A)} \sum_{c \in \text{range}(C)} \phi_1(a, \text{true}) \cdot \phi_2(c, \text{true}) \\
		&= \frac{1}{Z} \sum_{a \in \text{range}(A)} \phi_1(a, \text{true}) \sum_{c \in \text{range}(C)} \phi_2(c, \text{true}) \\
		&= \frac{1}{Z} \Bigg( \sum_{a \in \text{range}(A)} \phi_1(a, \text{true}) \Bigg)^2 \\
		&= \frac{1}{Z} \Bigg( \sum_{c \in \text{range}(C)} \phi_2(c, \text{true}) \Bigg)^2 \\
		&= \frac{1}{Z} \Big( \varphi_1 + \varphi_3 \Big)^2.
	\end{align*}
	This example illustrates the idea of using a representative of indistinguishable objects for computations (here, either $A$ or $C$ can be chosen as a representative for the group consisting of $A$ and $C$).
\end{example}

The idea of exploiting exponentiation can be generalised to groups consisting of $k$ indistinguishable objects to significantly reduce the computational effort for query answering.
To be able to exploit exponentiation during probabilistic inference, we need to ensure that the potential tables of factors within the same group are identical.
Indistinguishable objects frequently occur in many real world domains.
For example, in an epidemic domain, each person impacts the probability of having an epidemic equally.
That is, the probability of an epidemic depends on the number of sick people in the universe but is independent of which specific individual people are sick.

\section{Group Sizes of an Hierarchical Ordering}
\Cref{tab:level_groups222} shows for each level of the hierarchy in Fig. 2 of the main paper how many $\varepsilon$-equivalent groups of which size exist. 
Thus, it illustrates the increasing compression that is possible with increasing $\varepsilon$ values.

\begin{table}[htb] 
    \centering
    \begin{tabular}{c c c}
        \toprule
        Level & Number of total groups & Group Size (Frequency) \\ 
        \midrule
        0 & 10 & 1 (10),  \\ 
        1 & 9 & 2 (1), 1 (8)  \\ 
        2 & 8 & 2 (2), 1 (6) \\ 
        3 & 7 & 2 (3), 1 (4) \\ 
        4 & 6 & 4 (1), 2 (1), 1 (4) \\ 
        5 & 5 & 4 (1), 3 (1), 1 (3) \\ 
        6 & 4 & 4 (1) , 3 (1), 2 (1), 1 (1)\\ 
        7 & 3 & 7 (1), 2 (1), 1 (1)\\ 
        8 & 2 & 7 (1), 3 (1)\\ 
        9 & 1 & 10 (1)\\ 
        \bottomrule
    \end{tabular}
    \caption{Implicit group sizes for each level for given structure and pre-ordered \ac{fg} for Ex. 3 from the main paper.} 
    \label{tab:level_groups222}
\end{table}


\end{document}